\pdfoutput=1

\documentclass[11pt]{article}

\usepackage[final]{acl}

\usepackage{times}
\usepackage{latexsym}
\usepackage{color, colortbl}
\usepackage[T1]{fontenc}

\usepackage[utf8]{inputenc}

\usepackage{microtype}

\usepackage{inconsolata}

\usepackage{tcolorbox} %
\usepackage[dvipsnames]{xcolor}
\usepackage{graphicx,array}
\usepackage{enumitem} 
\usepackage{booktabs}
\usepackage{rotating}
\usepackage{multirow}
\usepackage{xspace}
\usepackage{soul}
\newcommand{\method}[0]{MDCure\xspace}
\newcommand{\methodrm}[0]{MDCureRM\xspace}

\def\mediumhline{\noalign{\hrule height.6pt}}

\title{\method: A Scalable Pipeline for Multi-Document Instruction-Following}

\author{
 \textbf{Gabrielle Kaili-May Liu}\textsuperscript{1}\quad
 \textbf{Bowen Shi}\textsuperscript{1}\quad
 \textbf{Avi Caciularu}\textsuperscript{2} \vspace{2pt}\\
 \textbf{Idan Szpektor}\textsuperscript{2}\qquad
 \textbf{Arman Cohan}\textsuperscript{1}
\\
\\
 \textsuperscript{1}Yale University
 \quad
 \textsuperscript{2}Google Research
\\
 \small{\texttt{\{kaili.liu, arman.cohan\}@yale.edu} $\quad$ 
 }
}

\begin{document}
\maketitle
\begin{abstract}
Multi-document (MD) processing is crucial for LLMs to handle real-world tasks such as summarization and question-answering across large sets of documents.
While LLMs have improved at processing long inputs, MD contexts still present unique difficulties, including management of inter-document dependencies, redundancy, and incoherent structures. 
To address this challenge, we introduce \method, a scalable and effective instruction data generation framework to enhance the MD capabilities of LLMs without the computational cost of pre-training or reliance on human-annotated data. \method generates high-quality synthetic MD instruction data over sets of articles via targeted prompts. We also introduce \methodrm, a cost-effective, MD-specific reward model to score and filter generated data based on their training utility for MD settings. MDCure is compatible with open- and closed-source models in addition to policy optimization methods such as PPO, enabling even small open-source models to surpass proprietary LLMs as strong generators of high-quality MD instruction data without further data filtering. With \method, we fine-tune a wide variety of LLMs up to 70B parameters in size from the FlanT5, Qwen2, and LLAMA3.1 model families. Extensive evaluations on a wide range of MD and long-context benchmarks spanning various tasks and domains show \method consistently improves performance over pre-trained baselines and base models by up to 75.1\%.

\end{abstract}

\section{Introduction}

With the rapid expansion of bodies of online texts in domains of science, finance, education, news, and more, multi-document (MD) processing becomes a critical aspect of LLMs' ability to understand and reason over large quantities of text information, with applications such as (query-focused) multi-document summarization (MDS) \citep{xiao-etal-2022-primera,giorgi-etal-2023-open}, multi-hop question-answering (MHQA) \citep{yang-etal-2018-hotpotqa}, and cross-document reasoning \citep{cattan2021scico}. 

 While LLMs can now process tens to hundreds of thousands of tokens, they struggle with the unique demands of multi-document understanding and reasoning: finding and aggregating information across diverse documents, resolving contradictions, handling redundant information, bridging information gaps, and constructing coherent narratives \citep{hosseini2024efficientsolutionsintriguingfailure, lior2024seamstochasticbenchmarkmultidocument,tang2024lciteeval,wang2024leave}. 
\begin{figure*}[h!]
\centering
\includegraphics[width=0.9\textwidth]{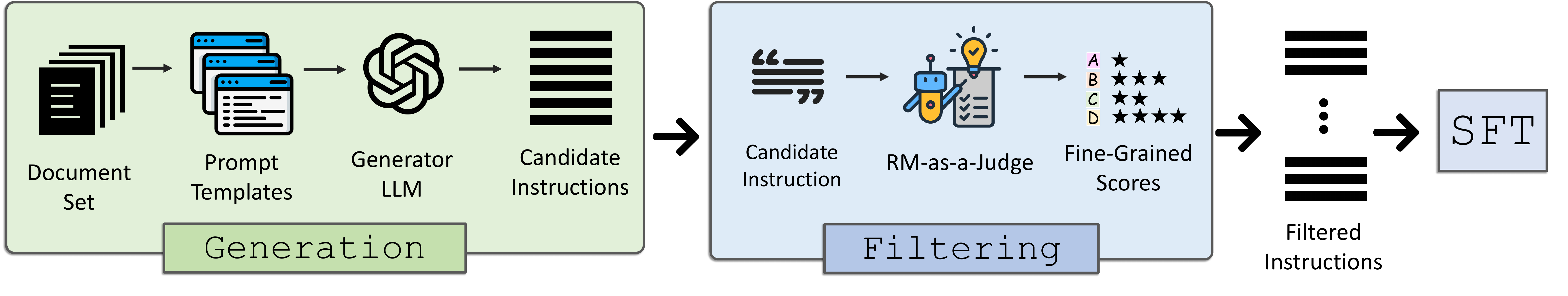}
\caption{
The MDCure pipeline generates diverse multi-document instructions, filters them using fine-grained scoring from MDCureRM, and tunes a base LLM to enhance its multi-document capabilities.
} \label{fig1}
\end{figure*}
Current approaches to improving MD capabilities primarily rely on continued pre-training of long-context models on related document sets \citep{caciularu-etal-2021-cdlm-cross, yasunaga-etal-2022-linkbert, caciularu-etal-2023-peek,peper-etal-2024-pelms}. However, such methods are not extensible to broader tasks 
and require large amounts of pre-training data.

We propose \method, an effective and scalable framework for {creating high-quality synthetic datasets designed to prioritize MD capabilities of LLMs and adapt LLMs for MD settings.}
\method builds upon recent work showing the value of small-scale, high-quality supervised data for efficiently improving LMs' downstream task performance \citep{zhou2023lima, sachdeva2024train, 10.5555/3666122.3668470}, in addition to addressing the challenges of data scarcity and complexity in MD tasks.
While curating organic, human-created training datasets might seem plausible, such approaches are often prohibitively expensive \citep{Cobbe2021TrainingVT, yue2024mammoth2scalinginstructionsweb} and suffer from limited scope, failing to capture complex reasoning patterns involved in MD settings \citep{srivastava2023imitationgamequantifyingextrapolating, chen2024essentialfactorscraftingeffective}.
Synthetic data leveraging frontier LLMs is becoming increasingly essential in post-training of open-source models \cite{abdin2024phi4technicalreport, lambert2024tulu3pushingfrontiers}. Our proposed \method is the first framework for creation of high-quality and MD-focused post-training datasets. Not only does \method achieve strong cross-task and cross-domain generalization in long-context MD settings, it also offers practical value due to its compatibility with both cost-effective commercial LLMs (e.g., GPT-3.5) and moderately-sized open-source LLMs (e.g., LLAMA3.1-70B).

\method divides the process to automatically \underline{cur}at\underline{e} high-quality MD instruction data into two phases. During \textit{Generation}, zero-shot prompt templates are used to create complex, cross-text instruction{-following} prompt-response pairs from a set of related documents provided as input context. During \textit{Filtering}, a multi-objective, MD-specific reward model, \methodrm, is trained to assess the generated instructions based on targeted criteria to ensure high-quality, diverse, MD-focused samples. We find that \methodrm is pivotal to improving the constitution of the resulting instruction dataset. Versus frontier LLMs, \methodrm overcomes limitations on the tradeoff between cost and quality, offering greater enhancements to MD data quality while also reducing filtering costs. Moreover, as we show, \methodrm seamlessly integrates with policy optimization methods such as PPO to iteratively improve synthetic MD instruction data generation policies, enabling even small open-source models such as LLAMA3.1-8B-Instruct to surpass SOTA proprietary models as strong generators of high-quality MD instruction data without further filtering or data curation.

We showcase the effectiveness of \method through extensive experiments on 6 MD and long-context benchmarks in over 6 content domains. To analyze the impact of \method at scale, we create \method instruction datasets of size 12K, 36K, and 72K to fine-tune FlanT5 \citep{chung2022scalinginstructionfinetunedlanguagemodels}, Qwen2-Instruct \citep{yang2024qwen2technicalreport}, and LLAMA3.1-Instruct \citep{dubey2024llama3herdmodels} models up to 70B parameters in size. Our results show that \method enables models to consistently achieve superior performance versus existing MD pre-training approaches, and to generalize well across diverse tasks and texts. Further, LLMs post-trained using our \method datasets consistently outperform prior instruction-following LLMs and strong long-context baselines, in both zero- and few-shot settings, with gains of up to \textbf{75.1\%} average improvement.
Our contributions are as follows:
\begin{itemize}[topsep=0pt, align=left, leftmargin=0pt, labelindent=6pt,
listparindent=\parindent, labelwidth=0pt, itemindent=!, itemsep=0pt, parsep=0pt]
    \item \method, a novel framework for obtaining high-quality, MD-focused post-training datasets to improve MD instruction-following capabilities of LLMs, which  achieves strong results with both open-source and proprietary cost-effective LLMs.
    \item \methodrm, a novel evaluator reward model specifically designed for the MD setting, to quality-filter MD instruction data more successfully and inexpensively versus proprietary LLMs.
    \item A suite of instruction data complementary to collections such as FLAN \citep{longpre2023flan} for improving MD task performance of any LLM.\footnote{We release our code at \url{https://github.com/yale-nlp/MDCure}.}
\end{itemize}

\section{Related Work}
\textbf{Multi-Document Modeling.} 
A common approach for MD modeling with LLMs is to perform flat concatenation of all input documents to process with a long-context model (e.g., Longformer \citep{beltagy2020longformerlongdocumenttransformer}, CoLT5 \citep{ainslie-etal-2023-colt5}). However, such models lack sufficient skill at synthesizing cross-document information \citep{caciularu-etal-2022-long,wolhandler-etal-2022-multi,chen2023walkingmemorymazecontext, hosseini2024efficientsolutionsintriguingfailure}. 
More recently, pre-training-based approaches have proven valuable and involve (continual) pre-training of long-context LMs on instances formulated over sets of related documents using masking and infilling objectives
\citep{pmlr-v119-zhang20ae,caciularu-etal-2021-cdlm-cross,xiao-etal-2022-primera,yasunaga-etal-2022-linkbert}. 
Other notable recent works leverage LLMs to better perform targeted content selection for MDS and QMDS \citep{caciularu-etal-2023-peek,kurisinkel2023controllablemultidocumentsummarizationcoverage, kurisinkel2023llmbasedmultidocumentsummarization, 10.1007/s11227-023-05457-z,peper-etal-2024-pelms}. In contrast, our \method is, to our knowledge, the first systematic framework for high-quality MD-specific instruction generation with LLMs to overcome the computational expense of pre-training, the limitations of heuristic-based data generation, and limited generalization of MD performance. 

\textbf{Synthetic Instruction Generation.}
Instruction tuning (IT) aligns LMs for diverse tasks via supervised instruction-answer pairs but is hindered by (1) the difficulty of obtaining high-quality instructions
\citep{song2024newpipelinegeneratinginstruction,ziegler2024craftdatasettaskspecificsynthetic}, and (2) unavailability of high-quality open-source instruction data \cite{lambert2024tulu3pushingfrontiers}.
To address this, recent works have focused on leveraging proprietary LLMs for synthetic data generation \citep{honovich-etal-2023-unnatural, wang-etal-2023-self-instruct, chen2024genqageneratingmillionsinstructions}. Most of these efforts target general \textit{single-document} tasks. A few, such as \citet{chen2024essentialfactorscraftingeffective, lupidi2024source2synthsyntheticdatageneration}, target MHQA by generating and merging single-hop and two-hop QA pairs. 
While effective, these approaches focus on QA without generalizing to broader MD tasks and rely on complex frameworks. In contrast, \method targets MD tasks broadly and enables cross-task generalization, while providing a simpler, more scalable solution to generate complex MD instruction data.
As synthetic data can be noisy, data selection using LLMs can be useful in filtering low quality data \citep{yue2024mammoth2scalinginstructionsweb, ziegler2024craftdatasettaskspecificsynthetic}. Recent approaches also leverage fine-grained rewards for task-specific data selection \citep{armorm}. 
Building on this, we introduce \methodrm, a multi-objective MD-specific reward model that refines instruction data effectively and cost-efficiently and which serves well both as a standalone judge and in RLAIF \citep{bai2022constitutionalaiharmlessnessai, lee2024rlaifvsrlhfscaling} settings.
Overall, with our \method framework, we construct and release open-source, high-quality instruction-following datasets that enables significant improvements on MD tasks. As we show, \method is not dependent on any specific LLM, works well with both open and proprietary models, and generalizes to many tasks.

\section{\method}
\method (Fig. \ref{fig1}) utilizes a two-step approach to produce high-quality MD instruction samples over sets of related documents. These samples can be used via IT to improve base LLMs’ MD abilities.

\subsection{Generation Phase} \label{generationphase}
During Generation, \method performs targeted construction of candidate instructions for a given set of documents by sampling from a set of carefully tailored zero-shot templates (App. \ref{genprompts})
to prompt a generator LLM. Adopting a principled approach, we design our prompt templates to encourage cross-document reasoning by requiring answers that synthesize information from multiple documents, with 
wide template variety ensuring diverse task formulations, ranging from single-word answers to detailed summaries. 
This yields synthetic MD data that reflects real-world complexity and encourages LLMs to leverage deep connections among texts to strengthen cross-document comprehension skills.
Versus \citet{caciularu-etal-2023-peek}, our instructions are more diverse and open-ended. Qualitative examples and diversity analysis of our \method instructions\footnote{Details regarding instruction format are in App. \ref{app:instructionformat}.} are provided in App.s \ref{app:sampledata} and \ref{app:diversity}, respectively. 
A systematic analysis of the effect of different prompt approaches on the quality of resulting generations is provided in \S\ref{analysis:prompteffect}.

\textbf{Dataset Source Selection.} 
\method's generation focuses on constructing candidate instructions from sets of documents, which may be sourced from existing datasets of related texts or created by clustering general text corpora. In our experiments, we use topically related document sets from the news domain to model cross-text relationships, building on prior work that shows training with related document sequences can enhance model performance \citep{caciularu-etal-2021-cdlm-cross, caciularu-etal-2023-peek, shi2023context}.
Nonetheless, our generation procedure is equally functional over sets of unrelated documents, which may be utilized to target creation of instruction samples concerning processing of conflicting or irrelevant cross-document information. We base our generated instructions on the NewSHead dataset \citep{headline2020}. Additional details are in App. \ref{newsheadappendix}.

\textbf{Generator Model. }\method is not generally dependent on any specific generator LLM. In combination with our proposed filtering procedure (\S \ref{filtering}), any strong instruction-following
LLM can be used to construct effective instruction data.
We utilize GPT-3.5-Turbo in experiments to balance generation quality and cost. We also show the compatibility and comparable efficacy of MDCure with use of open-source generator LLAMA3.1-70B in \S \ref{sec:opensourcegenerator}.\footnote{Other cost efficient LLMs such as GPT-4o-mini or Gemini 1.5 Flash and open-source LLMs such as LLAMA3.3-70B can also be used; our generation pipeline was done before the release of these more cost-efficient models.}

\subsection{Filtering Phase} \label{filtering}
\begin{figure}[t]
\centering
\includegraphics[width=\linewidth]{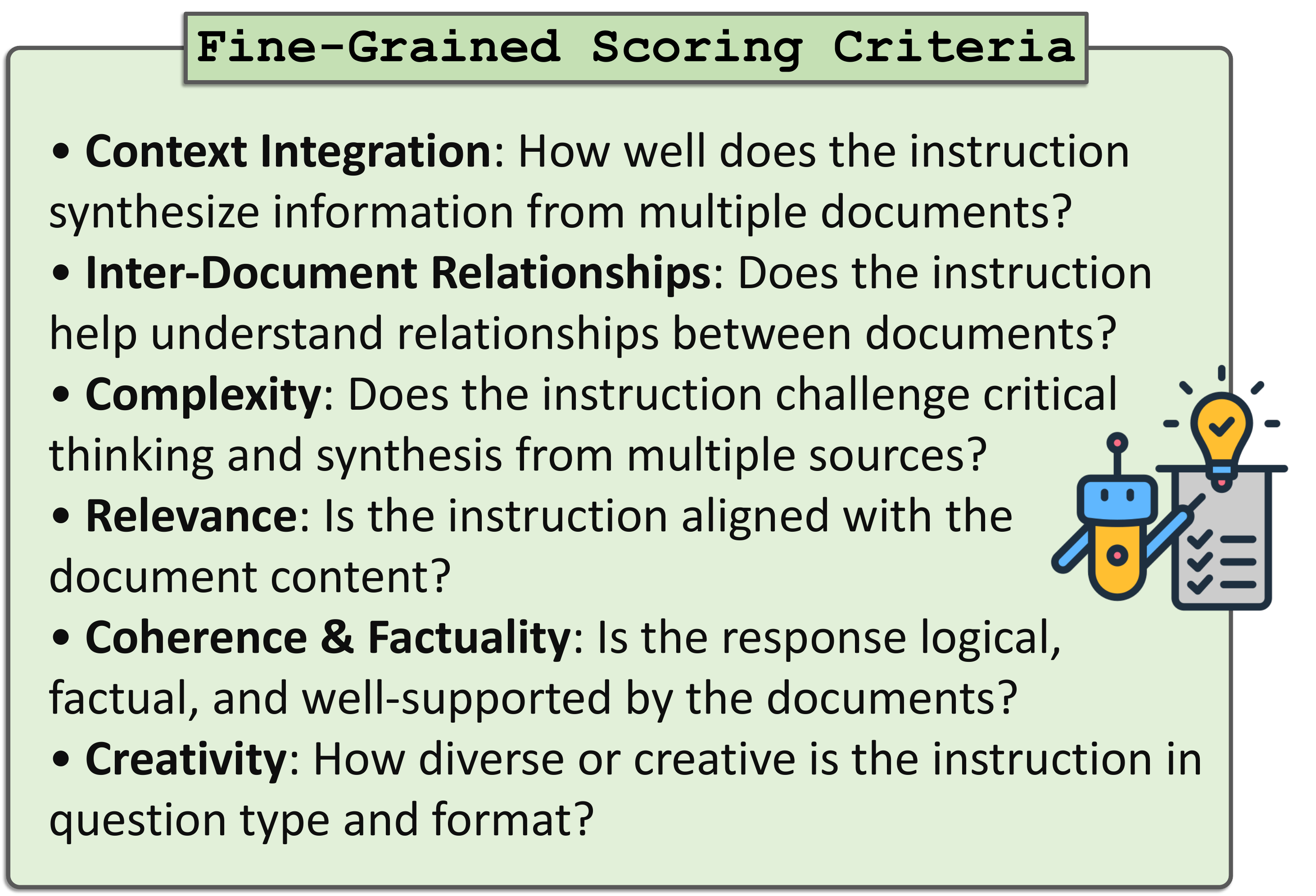}
\caption{Fine-grained scoring criteria utilized by MDCureRM for RM-as-a-Judge evaluation of candidate instruction quality in the Filtering phase of MDCure.} \label{fig3} %
\end{figure}

The generation prompts, while carefully designed, can still yield noisy or unsuitable instructions for MD settings. To address this, we apply a model-based filtering phase to remove low-quality, non-factual, or single-document instructions. Inspired by recent work on fine-grained RMs \citep{wu2023fine, wang2024arithmetic, armorm}, we train a fine-grained, \emph{MD-specific} reward model, \methodrm, to evaluate each instruction-answer pair based on six different criteria. These criteria capture both the overall quality of the instruction-response pairs and their effectiveness in handling multi-document content and are depicted in Fig. \ref{fig3}. This fine-grained RM-as-a-Judge mechanism refines the final curated instruction set to yield high-quality training data. We analyze the criticality of each \methodrm scoring criterion and the effect of their relative weightings via ablation studies in App. \ref{ablationsection}. A meta-evaluation to analyze the alignment of \methodrm with human annotations versus GPT-3.5-Turbo is provided in App. \ref{metaevaldeets}, where we observe agreements in line with LLM evaluation literature \citep{liu-etal-2024-benchmarking}. \S\ref{sec:ppo} further explores the efficacy of using reward signals provided by \methodrm to train a custom MD instruction generator model via PPO without the need for post-generation filtering.

\textbf{Training Setup.} We adopt a multi-objective reward modeling framework \citep{wu2023fine, wang2024arithmetic} to train \methodrm. Traditionally, reward models for LLM alignment are trained with Bradley-Terry loss \citep{Bradley1952RankAO} on pairwise data with binary preference annotations. However, we include the fine-grained information present in multi-objective ratings rather than binarizing the data, found to be effective in \citet{armorm}. Each training example for \methodrm consists of a prompt, response, and a 6-dimensional rating vector, with each dimension corresponding to a specific reward objective. We apply multi-objective regression to learn from these ratings, utilizing a regression loss function.

\textbf{Data.} Training data for \methodrm is obtained using an LLM-based pipeline without requiring human annotation. We first use GPT-4o-mini \citep{openai2024gpt4technicalreport} and Mistral-7B-Instruct-v0.2 \citep{jiang2023mistral} to generate MD instruction-answer pairs of varying quality.\footnote{We select these models due to their strong instruction-following capabilities, while considering the generation costs.} This results in roughly 20,000 data points. We then prompt GPT-4o to assign scores for each sample according to the six criteria, to be utilized as annotations to train the reward model. The specific prompt used to do so can be found in App. \ref{rmdataprompt}. The target reward scores are normalized to the range $[0, 1]$. 
Although using GPT-4o directly as the RM is also possible, it is prohibitively costly to scale. Training \methodrm ensures both cost-effectiveness and scalability for filtering large quantities of long-context data.

\textbf{Implementation.} \methodrm utilizes the Llama3-8B architecture, with weights initialized from a Bradley-Terry reward model \citep{xiong2024iterative} trained from Llama3-8B-Instruct \citep{dubey2024llama3herdmodels}.
The original output layer is replaced with a linear regression layer, outputting a 6-dimensional rating. The regression head processes the final-layer hidden states. We use Mean Squared Error (MSE) between target and predicted rewards as the loss function, freezing the base model during training, following \citet{armorm}. We discuss alternative architectures and the design of \methodrm in App. \ref{rmdesign}.

\textbf{Data Filtering.} 
During inference, \methodrm generates a 6-element rating for each candidate instruction. These scores are re-scaled to a 1-5 range and combined in a weighted average (details in App. \ref{scoreweight}). The top $N$ highest-scoring samples are selected as the final instructions.

\section{Experimental Setup}\label{expsetup}

We conduct extensive experiments to evaluate the efficacy of \method for improving multi-document modeling capabilities of LLMs. In particular, we provide evidence for the following:
\begin{itemize}[topsep=0pt, align=left, leftmargin=0pt, labelindent=6pt, listparindent=\parindent, labelwidth=0pt, itemindent=!, itemsep=0pt, parsep=0pt]
    \item The scalability of \method as we vary the size of the generated instruction data and the size of the base LLM being post-trained.
    \item The importance of MDCureRM filtering for curating high-quality MD instruction data.
    \item The strong cross-task and cross-domain generalization and superior results achieved by MDCure over pre-trained and long-context baselines.%
    \item The ability for \method data to excel over existing synthetic MD  \textit{pre-training} datasets.
    \item The adaptability of \method for few-shot and extended-context MD instruction data generation.
    \item The compatibility and efficacy of \method with open- and closed-source generator LLMs.
    \item The value of \methodrm toward providing reward signals for synthetic MD data policy optimization.
\end{itemize}

\textbf{Models and Training Details. }
We apply \method to LLMs across different widely used model families and sizes: FlanT5, 
Qwen2,
and LLAMA3.1.
We choose these models as they are capable open-source instruction-following models with varying architectures and parameter scales. We utilize the instruction-tuned variants as our base models to showcase the complementary effect of \method instructions versus typical IT data. To assess \method at scale, we create \method instruction datasets of size 12000, 36000, and 72000 examples\footnote{Dataset sizes were determined as a function of API costs and based on the sizes of representative IT datasets used in related works \citep{bai2024longalign, koksal2023longform}.} and use them to fine-tune FlanT5-Base and -Large (250M \& 750M), Qwen2-Instruct (1.5B \& 7B), and LLAMA3.1-Instruct (8B \& 70B). Hereafter, we refer to resulting models that have undergone \method post-training as ``\method'd'' models. To assess the efficacy of \methodrm, we additionally repeat these experiments using unfiltered and GPT-3.5-filtered  instruction sets.\footnote{We utilize GPT-3.5-Turbo for comparison as it is a more cost-effective alternative to commonly used scorer GPT4. Cheaper models such as GPT-4o-mini had not yet been released at the time of our starting experimentation.} Further data and training setup details are in App.s \ref{app:A} and \ref{trainingdeets}.

\textbf{Baselines.} Beyond their base model counterparts, our MDCure'd models are compared against state-of-the-art pre-trained models for multi-document processing, namely PRIMERA \citep{xiao-etal-2022-primera} and \textsc{QAMDen} \citep{caciularu-etal-2023-peek}, which are comparable in scale to FlanT5-Base and -Large, as well as several strong open-source long-context LLMs, namely LongAlign-7B \citep{bai2024longalign} {and ProLong-8B-64k-Instruct \citep{gao2024trainlongcontextlanguagemodels}}, which are comparable to Qwen2-7B-Instruct and LLAMA3.1-8B-Instruct, and Jamba-1.5B-Mini \citep{jamba} which has 54B parameters. We also compare MDCure'd models to SOTA general LLMs GPT-4o and Gemini 1.5 Pro \citep{gemini15report}. {Additionally, we inspect the effect of our instruction data versus baseline MD and long-context \textit{pre-training} data at the 7B- and 8B-scales by fine-tuning Qwen2-7B-Instruct with PRIMERA and \textsc{QAMDen} data and ProLong-8B-64k-Instruct with MDCure data (72k).} Further details regarding baselines are in App. \ref{baselinedeets}.

\textbf{Benchmarks.} We evaluate all baselines and MDCure'd models over a range of long- and short-form tasks through use of two challenging MD and long-context benchmarks: SEAM \citep{lior2024seamstochasticbenchmarkmultidocument} and ZeroScrolls \citep{zeroscrolls}. Within SEAM, we specifically consider the MultiNews, {OpenAsp,} MuSiQue, ECB+, and SciCo datasets to ensure coverage of a spread of domains and MD tasks.
We additionally profile 0-shot prompting performance over a suite of four widely used MD benchmarks: WikiHop \citep{welbl2018constructing} and HotpotQA-distractor \citep{yang-etal-2018-hotpotqa} for MHQA,
Multi-XScience \citep{lu-etal-2020-multi-xscience} for MDS, and QMDSCNN \citep{Pasunuru2021DataAF} for QMDS. While these datasets were built for use in fine-tuning evaluation, we draw upon the prompting approach proposed by \citet{zeroscrolls} to formulate effective 0-shot prompts for each model. App.s \ref{benchmarkdeets}  and \ref{evaldeets} provide additional details.

\textbf{Metrics.} For SEAM and ZeroScrolls
we employ the metrics and corresponding implementations defined by each for their component datasets. For WikiHop and HotpotQA-distractor, we use the standard F1 score.
For Multi-XScience and QMDSCNN, we utilize LLM scores issued by GPT-3.5-Turbo and Gemini 1.5 Flash, respectively, to assess summarization coherence, relevance, fluency, and consistency. This follows recent adoption of LLM-as-a-Judge as a proxy for human-level rating \citep{liu2024reife, ye2024flaskfinegrainedlanguagemodel}, given the limitations of traditional metrics like ROUGE for long-form generation quality assessment. \citep{10.1609/aaai.v33i01.33016220, gao2024prolong}. Further details regarding the evaluator LLMs and the prompts are in App. \ref{llmevaldeets}.

\definecolor{Gray}{gray}{0.9}
\definecolor{Gray2}{gray}{0.95}
\definecolor{OurColor}{RGB}{255, 241, 214}
\definecolor{BaselineColor}{RGB}{219, 233, 255}

\newcommand{\hlcolorone}[1]{\sethlcolor{BaselineColor}\hl{#1}}
\newcommand{\hlcolortwo}[1]{\sethlcolor{OurColor}\hl{#1}}

\section{Results}
\begin{table*}[h!]
\centering
\footnotesize\setlength{\tabcolsep}{4pt}
\begin{tabular}{@{}cc|l*{4}{c}|*{6}{c}|*{1}{c}|*{1}{c}@{}}
\toprule
&  & \multirow{2}{*}{\parbox[l]{1.2cm}{\centering Model / IT Data Setting}} & \multirow{2}{*}{HQA} & \multirow{2}{*}{WH} & \multirow{2}{*}{MXS} & \multirow{2}{*}{QC} & \multicolumn{6}{c}{SEAM} & \multirow{2}{*}{ZS Score} & \multirow{2}{*}{\centering Avg} \\
\cmidrule(lr){8-13} 
& & & & & & & ECB+ & MN & MSQ & OA&SC & Avg  & & \\
\midrule
\multirow{8}{*}{\rotatebox[origin=c]{90}{\parbox[c]{3.8cm}{\centering \textsc{FlanT5}}}} & \multirow{4}{*}{\rotatebox[origin=c]{90}{\parbox[c]{2.1cm}{\centering Base}}} 
& \cellcolor{BaselineColor}PRIMERA  &\cellcolor{BaselineColor} 0.4 & \cellcolor{BaselineColor}0.5 & \cellcolor{BaselineColor}70.7	&\cellcolor{BaselineColor}	24.2 &\cellcolor{BaselineColor}7.1	&\cellcolor{BaselineColor}7.7	&\cellcolor{BaselineColor}0.5	&\cellcolor{BaselineColor}3.9&\cellcolor{BaselineColor}10.4	&\cellcolor{BaselineColor}\textbf{5.9}&\cellcolor{BaselineColor}4.2&\cellcolor{BaselineColor}8.8\\
& & \cellcolor{BaselineColor}\textsc{QAMDen} &\cellcolor{BaselineColor} 1.8 &\cellcolor{BaselineColor} 1.9 &\cellcolor{BaselineColor} 63.6	&\cellcolor{BaselineColor}	27.1 &\cellcolor{BaselineColor} 0.0	&\cellcolor{BaselineColor}0.4	&\cellcolor{BaselineColor}0.0	&\cellcolor{BaselineColor}1.0&\cellcolor{BaselineColor}3.1	&\cellcolor{BaselineColor}0.9 &\cellcolor{BaselineColor}3.8&\cellcolor{BaselineColor}7.2\\
& & None & 4.4 & 45.1 & 38.7 & 48.0 & 0.0	& 5.8	& 0.2	&2.6& 0.1	& 1.7& 13.1&14.5\\
& & Unfiltered & 31.7 & 47.5 & 89.8	& 52.2 & 0.0	&6.7&	0.1&2.7&	0.2&	1.9 & 21.0&23.2\\
& & GPT-Filtered & 44.1 & 46.0 & 92.4	&	54.2 & 0.0&	6.0	&0.2&2.7&	0.0&	1.8 &21.3& 24.1\\
& & \cellcolor{OurColor} \methodrm &\cellcolor{OurColor} \textbf{47.3} &\cellcolor{OurColor} \textbf{48.3} &\cellcolor{OurColor} \textbf{93.8}	&\cellcolor{OurColor}	\textbf{57.3} &\cellcolor{OurColor} 0.0&\cellcolor{OurColor}	7.5	&\cellcolor{OurColor}0.3&\cellcolor{OurColor}2.7	&\cellcolor{OurColor}0.0	&\cellcolor{OurColor}2.1 &\cellcolor{OurColor} 	\textbf{22.6}	&\cellcolor{OurColor}\textbf{25.4}\\
\cmidrule[0.7pt]{2-15}
& \multirow{4}{*}{\rotatebox[origin=c]{90}{\parbox[c]{1cm}{\centering Large}}} 
& None & 24.4 & 54.6 & 70.9	&	61.8 & 1.1	&7.9	&0.1	&3.1&0.3	&2.5& 23.2&24.0\\
& & Unfiltered & 46.9 & 62.9 & 91.1	&	64.7& 0.7&	8.0	&0.0&3.8	&0.1	&2.5 &24.2&27.3\\
& & GPT-Filtered & 46.3 & 65.1 & 91.8	&	64.0 & 0.5&	8.3	&0.0&3.9&	0.1	&2.6 &24.2&27.5\\
& & \cellcolor{OurColor} \methodrm &\cellcolor{OurColor} \textbf{49.6} &\cellcolor{OurColor} \textbf{66.1} &\cellcolor{OurColor} \textbf{93.1}	&\cellcolor{OurColor}	\textbf{66.0} &\cellcolor{OurColor} 1.2&\cellcolor{OurColor}	9.1	&\cellcolor{OurColor}0.1	&\cellcolor{OurColor}4.2&\cellcolor{OurColor}0.0&\cellcolor{OurColor}	\textbf{2.9} &\cellcolor{OurColor} \textbf{25.3}	&\cellcolor{OurColor}\textbf{28.5}\\
\cmidrule[0.7pt]{1-15}
\multirow{8}{*}{\rotatebox[origin=l]{90}{\parbox[c]{3.3cm}{\centering \textsc{Qwen2-Ins}}}} & \multirow{4}{*}{\rotatebox[origin=c]{90}{\parbox[c]{1cm}{\centering 1.5B}}} 
& None & 21.5 & 17.8 & 93.3	&	73.3 & 9.4	& 10.6& 	0.5	& 5.0&13.0	& 7.7 &24.0&25.5\\
& & Unfiltered & 32.9 & 30.5 & 94.2	&	79.3 &15.7	& 12.0& 	0.4	& 5.0&16.8	& 10.0 &23.9&27.7\\
& & GPT-Filtered & 33.3 & 32.9 & 94.2	&	81.3 & 16.1	&12.0	&0.5	&5.1&15.9&	9.9 	&24.2&28.1\\
& & \cellcolor{OurColor}\methodrm &\cellcolor{OurColor} \textbf{37.7} &\cellcolor{OurColor} \textbf{34.8} &\cellcolor{OurColor} \textbf{94.4}	&\cellcolor{OurColor}	\textbf{82.9} &\cellcolor{OurColor} 16.4	&\cellcolor{OurColor} 12.0	&\cellcolor{OurColor} 0.6&\cellcolor{OurColor} 5.9	&\cellcolor{OurColor}18.1	&\cellcolor{OurColor} \textbf{10.6} &\cellcolor{OurColor}\textbf{25.6}	&\cellcolor{OurColor}\textbf{29.4}\\
\cmidrule[0.7pt]{2-15}
& \multirow{4}{*}{\rotatebox[origin=c]{90}{\parbox[c]{1.7cm}{\centering 7B}}} 
 & \cellcolor{BaselineColor}LongAlign-7B &\cellcolor{BaselineColor} 10.4 &\cellcolor{BaselineColor} 14.3 &\cellcolor{BaselineColor} 92.2	&\cellcolor{BaselineColor}	83.3 &\cellcolor{BaselineColor} 11.5&\cellcolor{BaselineColor}	16.5	&\cellcolor{BaselineColor}0.0&\cellcolor{BaselineColor}4.1	&\cellcolor{BaselineColor}16.8	&\cellcolor{BaselineColor}9.8 &\cellcolor{BaselineColor}23.2&\cellcolor{BaselineColor}25.3\\
& & None & 30.5 & 39.6 & 95.6	&	79.3 & 5.0	& 11.9	& 0.5	&6.4 &13.1	& 7.4 &23.9&27.4\\
& & Unfiltered & 40.5 & 43.3 & 94.7	&	84.2 & 8.0	&15.3	&0.5	&6.6&11.9&	8.5&26.3& 29.9\\
& & GPT-Filtered & 42.0 & 44.0 & 94.7	&	85.3 & 8.7	&15.3	&0.9	&6.6&12.1	&8.7 &28.7& 31.4\\
& & \cellcolor{OurColor} \methodrm &\cellcolor{OurColor} \textbf{44.7} &\cellcolor{OurColor} \textbf{46.0} &\cellcolor{OurColor} \textbf{95.1}	&\cellcolor{OurColor}	\textbf{87.3} &\cellcolor{OurColor} 13.8	&\cellcolor{OurColor}15.4	&\cellcolor{OurColor}0.6	&\cellcolor{OurColor}6.7&\cellcolor{OurColor}14.9&\cellcolor{OurColor}	\textbf{10.3} &\cellcolor{OurColor} \textbf{29.8}	&\cellcolor{OurColor}\textbf{32.7}\\
\cmidrule[0.7pt]{1-15}
\multirow{6}{*}{\rotatebox[origin=l]{90}{\parbox[c]{3.7cm}{\centering \textsc{Llama3.1-Ins}}}} & \multirow{4}{*}{\rotatebox[origin=c]{90}{\parbox[c]{1.7cm}{\centering 8B}}} 
 & \cellcolor{BaselineColor}ProLong-8B &\cellcolor{BaselineColor} 43.6 &\cellcolor{BaselineColor} 34.3 &\cellcolor{BaselineColor} 85.8	&\cellcolor{BaselineColor}	54.7 &\cellcolor{BaselineColor}9.5 &\cellcolor{BaselineColor}	17.4	 &\cellcolor{BaselineColor}0.4	 &\cellcolor{BaselineColor}10.7	 &\cellcolor{BaselineColor}18.0	 &\cellcolor{BaselineColor}11.2  &\cellcolor{BaselineColor}32.2&\cellcolor{BaselineColor}32.1\\
& & None & 35.5 & 27.1 & 95.1	&	65.3 & 10.5&	15.0	&0.6&7.6&	17.4	&10.2 & 18.7& 24.3\\
& & Unfiltered & 37.6 & 34.4 & 84.7	&	90.4 & 15.3&	16.3	&0.5&7.8&	18.9&	11.8 &28.6&31.1\\
& & GPT-Filtered & 38.0 & 42.3 & 95.3	&	87.8 & 8.7	&16.0	&0.6	&7.8&18.3	&10.3&29.6&32.1\\
& & \cellcolor{OurColor}\methodrm &\cellcolor{OurColor} \textbf{44.7} &\cellcolor{OurColor} \textbf{43.7} &\cellcolor{OurColor} \textbf{95.3}	&\cellcolor{OurColor}	\textbf{93.8} &\cellcolor{OurColor} 16.3	&\cellcolor{OurColor}16.4&\cellcolor{OurColor}	0.6	&\cellcolor{OurColor}7.9&\cellcolor{OurColor}18.5	&\cellcolor{OurColor}\textbf{11.9}&\cellcolor{OurColor}	\textbf{30.9}	&\cellcolor{OurColor}\textbf{34.0}\\
\cmidrule[0.7pt]{2-15}
& \multirow{2}{*}{\rotatebox[origin=c]{90}{\parbox[c]{1.7cm}{\centering 70B}}} 
 & \cellcolor{BaselineColor}Jamba 1.5 Mini &\cellcolor{BaselineColor} 47.5 &\cellcolor{BaselineColor} 41.8 &\cellcolor{BaselineColor} 94.2	&\cellcolor{BaselineColor}	87.1 &\cellcolor{BaselineColor} 20.1	&\cellcolor{BaselineColor}14.2	&\cellcolor{BaselineColor}0.0	&\cellcolor{BaselineColor}5.3&\cellcolor{BaselineColor}20.4	&\cellcolor{BaselineColor}12.0 &\cellcolor{BaselineColor}34.1&\cellcolor{BaselineColor} 35.3\\
 &&  None & 53.9 & 38.1 & 95.1	&	88.2 & 25.1	& 21.9& 	0.6	&6.1& 11.5	& 13.0 &36.4&37.1\\
& & Unfiltered & 55.9 & 40.6 & 88.7	&	70.4 &3.6	&21.9&	0.6	&6.3	&11.2&	8.7 &34.9&34.1\\
& & GPT-Filtered & 57.4 & 41.2 & 88.2	&	74.9 & 4.9	&22.0	&0.7&	6.4	&10.7&	8.9&37.5&35.9\\
& &  \cellcolor{OurColor} \methodrm &\cellcolor{OurColor} \textbf{58.4 }&\cellcolor{OurColor} \textbf{45.5} &\cellcolor{OurColor} \textbf{95.1}	&\cellcolor{OurColor}	\textbf{88.7} &\cellcolor{OurColor} 25.2	&\cellcolor{OurColor} 22.1&\cellcolor{OurColor} 	0.7	&\cellcolor{OurColor}6.7&\cellcolor{OurColor} 12.0	&\cellcolor{OurColor}\textbf{ 13.3}&\cellcolor{OurColor}	\textbf{37.7}	&\cellcolor{OurColor}\textbf{38.5}\\
\cmidrule[0.7pt]{1-15}
& & GPT-4o & 57.5 & 50.0 & \textbf{100.0}	&	\textbf{94.4} & 8.9	 &18.6 &	0.3 &14.2&	28.9	 &14.1 &	\textbf{39.8}	&\textbf{40.6}\\
& & Gemini 1.5 Pro & \textbf{66.6} & \textbf{55.6} & 93.8	&	79.8 & 21.4&	17.8	&0.6	&15.1&30.1	&\textbf{17.0}&32.0  &36.9\\
\bottomrule
\end{tabular}
\caption{Evaluation of MDCure'd models versus baselines on 6 benchmarks in the zero-shot prompting setting. The rightmost ``Avg'' column reports the average of individual dataset scores. Dataset abbreviations are described in App. \ref{abbreviations}. Full results on ZeroScrolls are provided in App. \ref{fullresultsappendix}. Rows specified by ``\methodrm'' refer to our full \method pipeline applied to the corresponding base model and size. Size-comparable baselines are highlighted in \hlcolorone{blue} and results of our final method is highlighted in \hlcolortwo{yellow}. Bold numbers show best performance in each group.}\label{maintable} 
\vspace{-2.5mm}
\end{table*}
\subsection{Main Results} 

We report experimental results in Table \ref{maintable}{, with full results on ZeroScrolls included in App. \ref{fullresultsappendix}}. For fair comparison, we report all \method results based on training with 72K samples, which led to best performance overall across all models. 
Our key findings are summarized as follows.

\textbf{MDCure instructions are effective across model families and sizes.} MDCure exhibits the best performance across all base models, demonstrating a clear and consistent advantage relative to non-MDCure’d base models over all benchmarks. In particular, we see improvements of up to 975\%, 96\%, 143\%, 434\%, 39\%, and 73\% for HotpotQA, WikiHop, Multi-XScience, QMDSCNN, SEAM, and ZeroScrolls, respectively. MDCure’d LLAMA3.1-70B-Instruct achieves the best results across benchmarks, followed by MDCure’d Qwen2-7B-Instruct. As our models are trained from -instruct versions of base LLMs, it is clear that MDCure instruction data imparts complementary abilities versus typical IT data.

The relative improvement of MDCure’d models versus corresponding base models varies across model families, and improvements tend to shrink with model size. For FlanT5, MDCure provides 75.1\% average improvement at the 250M-parameter scale (FlanT5-Base) versus 18.9\% at the 750M-parameter scale (FlanT5-Large). For LLAMA3.1, MDCure provides 40.2\% average improvement at the 8B scale versus 3.8\% at the 70B scale. For Qwen2, {we observe alternatively that} MDCure gains shift from 15.4\% on average at the 1.5B scale to 19.4\% at the 7B scale. More generally, these results suggest the importance of instruction quality may weaken with base model scale and capability, similar to findings by \citet{ivison2023camels}.

\textbf{Filtering with \methodrm is essential to improve MD capabilities of LLMs.} Comparing to models trained with unfiltered MDCure instruction samples, MDCure’d models trained with the highest-scoring \methodrm-filtered instructions are superior across all benchmarks and base models. Furthermore, MDCureRM outperforms GPT-3.5-Turbo as a judge of MD data quality and efficacy, evidenced by MDCure'd models superiority over those trained with GPT-filtered instructions. As use of naive, unfiltered instructions yields relatively little improvement to MD capabilities especially at larger base model scales, this suggests curation of high-quality synthetic MD data is a nontrivial goal which \method is able to efficaciously accomplish.

\textbf{MDCure improves performance on both MD and single-document long-context tasks, enabling cross-task and cross-domain generalization to unseen tasks and domains.} Across MD benchmarks, MDCure enables models to achieve improved performance in all settings, with strong generalization to out-of-distribution (OOD) tasks such as MD coreference resolution, MD classification, text reordering, and aggregated sentiment classification and to OOD domains such as science, literature, and media. Beyond MD tasks, \method generally boosts single-document long-context performance, particularly on tasks in ZeroScrolls, which emphasize long-context abilities.
These trends are consistent across downstream task domains, suggesting \method data boosts MD performance in a domain-general way. 

\begin{table}[t]
\centering\footnotesize\setlength{\tabcolsep}{2.7pt}
\begin{tabular}{@{}lccccccc@{}}
\toprule
Model               & HQA  & WH   & MXS  & QC   &  SEAM & ZSS & Avg \\ \midrule
ProLong (PL)          & 43.6 & 34.3 & 85.8	&	54.7 &  11.2 & 33.6 &  32.1      \\
PL+MDCure & \textbf{45.8} & \textbf{45.1} & \textbf{87.3}	&	\textbf{75.6} &  \textbf{12.5} &\textbf{34.7} & \textbf{34.9}  \\ \bottomrule
\end{tabular}
\caption{\method performance when applied to an already strong long-context LLM, ProLong-8B.}
\label{tab:prolong-comparison}
\end{table}
\begin{table}%
\centering 
\renewcommand{\arraystretch}{1.1}
\footnotesize\setlength{\tabcolsep}{2.5pt}
\begin{tabular}{@{}lccccccc@{}}\toprule
Data Source  & HQA & WH & MXS & QC &  SEAM  & ZSS & Avg\\\hline
Qwen2-7B-Ins & 30.5 & 39.6 & 95.6	&	79.3 & 7.4 & 23.9 & 27.4 \\ 
+\method & \textbf{44.7} & \textbf{46.0} & \textbf{95.1}	&	\textbf{87.3} &	\textbf{10.3} & \textbf{29.8}	&\textbf{32.7}\\
+PRIMERA & 44.6 & 45.4 & 88.0	&	48.2	&8.1 &27.9&28.7\\
+\textsc{QAMDen} & 42.6 & 45.8 & 87.1	&	52.7 &	8.8 & 27.2&28.6\\
\bottomrule
\end{tabular} 
\caption{Efficacy of \method data versus synthetic MD pre-training data, applied to Qwen2-7B-Instruct.}\label{analysis:datasource}
\vspace{-2mm}
\end{table}

\textbf{\method consistently outperforms pre-trained long-context baselines.} Our MDCure’d models achieve results that exceed the performance of size-comparable baselines\footnote{We consider Jamba 1.5 Mini (54B) comparable with LLAMA3.1-70B-Instruct, as their parameter scales are within the same ballpark, falling within the same order of magnitude.} trained on much longer inputs (64-256K tokens) for all benchmarks by a substantial margin, confirming the efficacy of our design. {For a controlled setup, we apply \method to strong open-source long-context LLM Prolong-8B \citep{gao2024prolong} to study if our data provides additional advantages to long-context continual pre-training and fine-tuning. Table \ref{tab:prolong-comparison} demonstrates the results. Note that Prolong-8B has undergone extensive training on $\sim$81B tokens to improve long-context performance. Fine-tuning ProLong-8B with much less MDCure data ($\sim$162M tokens) yields further improvement, verifying the added long-context value of our data.}

\textbf{Our \method pipeline improves over existing MD synthetic pre-training data.}
{We also compare our \method pipeline with existing MD-focused synthetic pre-training datasets inluding PRIMERA and \textsc{QAMDen}. }
As indicated in Table \ref{analysis:datasource},
MDCure data offers performance improvements across benchmarks at the 7B-scale.
The weak performance of the original \textsc{PRIMERA} and \textsc{QAMDen} models (Table \ref{maintable})
can be explained as they are intended to serve as strong base pre-trained models for finetuning as opposed to zero-shot prompting settings. Overall, our results emphasize the utility of our framework as a powerful approach for improving MD abilities of LLMs via post-training.

\subsection{Influence of Data Scale} \label{analysis:data_scale}
\begin{table}%
\centering 
\renewcommand{\arraystretch}{1.1}
\footnotesize\setlength{\tabcolsep}{3pt}
\begin{tabular}{@{}lcccccc@{}}\toprule
\# IT Samples  & HQA & WH & MXS & QC &  SEAM  & ZSS \\\hline\rowcolor{white} \multicolumn{7}{c}{FlanT5-Base}\\\hline
None & 4.4 & 45.1 & 38.7	&	48.0 & 1.7 & 13.1 \\ 
12K RM-Filtered & 43.8 & \textbf{48.4} & 92.4 & 54.0 & 1.9 & 22.0 \\  
36K RM-Filtered  & 45.7 & 46.1 & 92.7 & 56.7 & 1.9 & 22.2 \\
72K RM-Filtered & \textbf{47.3} & 48.3 & \textbf{93.8}	&	\textbf{57.3} & \textbf{2.1} &  \textbf{22.6} \\ 
72K Unfiltered & 31.7 & 47.5 & 89.8 & 52.2 & 1.9 &  21.0 \\  \hline
  \rowcolor{white}\multicolumn{7}{c}{Qwen2-7B-Instruct}\\\hline
None & 30.5 & 39.6 & 95.6	&	79.3 & 7.4 & 23.9 \\  
12K RM-Filtered & 44.1 & 45.4 & 94.7 & 82.7 & 8.9 & 25.9 \\  
36K RM-Filtered  & 44.4 & 45.8 & 94.7 & 84.7 & 8.6 & 28.1 \\
72K RM-Filtered & \textbf{44.7} & \textbf{46.0} & \textbf{95.1}	&	\textbf{87.3}  & \textbf{10.3} & \textbf{29.8} \\ 
72K Unfiltered & 40.5 & 43.3 & 94.7 & 82.2 & 8.5 & 26.3  \\ 
\bottomrule
\end{tabular} 
\caption{Performance at different training data scales.}\label{data_scale}
\vspace{-2mm}
\end{table}
To understand the importance of data quantity, we compare the impact of fine-tuning with \method instruction datasets of various sizes. We utilize FlanT5-Base and Qwen2-7B-Instruct as base models to demonstrate the effect of data size at different model scales. Results are compared versus non-MDCure’d baselines and shown in Table \ref{data_scale}. We observe consistent, albeit modest, improvements to downstream MD performance as we scale up the instruction data size.
Notably, use of 12K \method-filtered samples consistently surpasses performance with use of 72K \textit{unfiltered} samples for both base models, consistent with other work finding use of only a few thousand high-quality samples is sufficient for alignment \citep{zhou2023lima}. This evidences \methodrm's ability to promote data efficiency and reduce costs involved in generating and using an effective MD IT dataset.

\begin{table}
\renewcommand{\arraystretch}{1.1}
\centering\footnotesize\setlength{\tabcolsep}{1.7pt}
\begin{tabular}{@{}lcccccc@{}}\toprule
  & HQA & WH & MXS & QC &  SEAM & ZSS \\\hline
   \rowcolor{white} \multicolumn{7}{c}{Zero-Shot Prompting Evaluation}\\\hline
LLAMA3.1-8B-Instruct & 35.5 & 27.1 & 95.1 & 65.3 & 10.2 & 18.7 \\  
+\method & \textbf{41.0} & 40.2 & 95.6 & 87.6 & 12.0 & 31.7 \\  
+\method\hspace{-2pt}+LongContext & 38.9 & \textbf{40.5} & \textbf{95.8} & \textbf{88.2} & \textbf{12.1} & \textbf{32.5} \\  \hline
   \rowcolor{white} \multicolumn{7}{c}{5-Shot Prompting Evaluation}\\\hline
LLAMA3.1-8B-Instruct & 57.5 & 44.2 & 88.2 & 65.1 & 15.1 & ---\\  
+\method & 63.9 & 47.0  & 89.8 & \textbf{73.8} & 17.3 & --- \\  
+\method+Few-Shot & \textbf{65.0} & \textbf{49.2} & \textbf{90.7} & \textbf{73.8} & \textbf{17.9} & --- \\  %
    \bottomrule
\end{tabular} 
\caption{Efficacy of \method for few-shot and long-context MD instruction data generation.}\label{fs_long_context}
\vspace{-2mm}
\end{table}

\subsection{Generalization to Few-Shot \& Extended-Context Training} \label{sec_fs_32k}
We investigate the scalability of \method instructions to the few-shot prompting and extended-context settings. The few-shot setting is of interest as it provides an inexpensive alternative for adapting models to new tasks with limited compute \citep{Dong2023ASF}. Likewise, real-world MD tasks often occur over ``extreme’’ numbers of documents such as in extreme MDS \citep{lu-etal-2020-multi-xscience}. We extend our \method data to these settings 
by packing few-shot exemplars or distractor articles into instruction inputs up to 32K tokens in length.\footnote{Details of extended-context data construction is in App. \ref{longcontextdata}. Context length is selected based on GPU memory limits.} 
Experiments utilize LLAMA3.1-8B-Instruct as the base model, with training setup details described in App. \ref{trainingdeets}. Results are displayed in Table \ref{fs_long_context}, where we evaluate the few-shot setting via 5-shot prompting, omitting ZeroScrolls which targets 0-shot evaluation. We find that use of \method with extended contexts further improves performance for most benchmarks. Combined with prior results showing \method generalizes beyond training context length, this suggests \method is effective and applicable in both resource-rich and resource-constrained settings, enhancing its practicality.

\subsection{Efficacy with Open-Source Generation} \label{sec:opensourcegenerator}
We demonstrate the distinct cost advantage of MDCure over existing synthetic data pipelines which rely on GPT-based generations and target only non-MD contexts by studying the impact of using an open-source generator model in place of GPT-3.5-Turbo. Following the same procedure as in previous settings, we alternately use LLAMA3.1-70B-Instruct generation with \methodrm filtering to create an MDCure instruction dataset of size 12K examples, since \S \ref{analysis:data_scale} finds this quantity to be sufficient for strong improvements to MD task performance.  Fine-tuning results on Qwen2-1.5B-Instruct, Qwen2-7B-Instruct, and LLAMA3.1-8B-Instruct are reported in Table \ref{tab:llamageneration}. Across model sizes and families, use of open-source generations in fact leads to better performance versus use of GPT generations. These emphasize the flexibility of \method and validate the unique cost-saving nature of our synthetic data framework. 

\begin{table}
\centering 
\renewcommand{\arraystretch}{1.1}
\footnotesize\setlength{\tabcolsep}{2.2pt}
\begin{tabular}{@{}lccccccc@{}}\toprule
Generator LLM  & HQA & WH & MXS & QC &  SEAM  & ZSS & Avg \\\hline\rowcolor{white} \multicolumn{8}{c}{Qwen2-1.5B-Instruct}\\\hline
GPT-3.5-Turbo  & \textbf{37.0}&	33.2	&\textbf{94.4}	&	\textbf{83.3}&	10.0	&24.4&	42.3 \\
LLAMA3.1-70B & 36.7	&\textbf{33.9}&	94.2	&	82.4	&\textbf{10.3}	&\textbf{24.5}	&\textbf{42.8} \\  \hline
  \rowcolor{white}\multicolumn{8}{c}{Qwen2-7B-Instruct}\\\hline
GPT-3.5-Turbo  & 44.1	&45.4&	94.7	&	82.7	&\textbf{8.9}	&25.9	&43.0\\
LLAMA3.1-70B & \textbf{46.9}&	\textbf{45.6}	&\textbf{94.9}	&	\textbf{82.7}	&8.2	&\textbf{28.7}	&\textbf{44.0} \\  \hline
  \rowcolor{white}\multicolumn{8}{c}{LLAMA3.1-8B-Instruct}\\\hline
GPT-3.5-Turbo  & \textbf{40.9}	&\textbf{41.6}	&95.8	&	87.6	&11.9	&30.7&	50.3\\
LLAMA3.1-70B & 40.0	&37.5&	\textbf{97.3}	&	\textbf{88.0}&	\textbf{12.2}	&\textbf{34.3}	&\textbf{52.9} \\  
\bottomrule
\end{tabular} 
\caption{\method results with different generator LLMs.}\label{tab:llamageneration}
\vspace{-2mm}
\end{table}

\subsection{Integration with Policy Optimization} \label{sec:ppo}

Observing MDCureRM’s strength as a judge of MD instruction quality, we explore the value of using \methodrm with reinforcement learning methods such as PPO to iteratively improve synthetic MD data generations. We utilize LLAMA3.1-8B-Instruct as the initial policy model\footnote{We select this model as it is relatively small while remaining effective at instruction-following.} and use \methodrm to provide reward signals whilst training with a PPO objective. Details of our PPO training setup can be found in App. \ref{app:ppo}. As in \S\ref{sec:opensourcegenerator}, we use the final policy model to create a size-12K \method instruction dataset and report fine-tuning results on Qwen2-1.5B-Instruct, Qwen2-7B-Instruct, and LLAMA3.1-8B-Instruct in Table \ref{tab:ppotable}. Across benchmarks, use of generations from PPO-trained LLAMA3.1-8B-Instruct enables all three models to achieve close or superior performance versus use of GPT generations. Notably, \method enables models far smaller than SOTA LLMs to become effective MD instruction generators, further enhancing the cost-effectiveness of our approach.
\begin{table}
\centering 
\renewcommand{\arraystretch}{1.1}
\footnotesize\setlength{\tabcolsep}{2pt}
\begin{tabular}{@{}lccccccc@{}}\toprule
Generator LLM  & HQA & WH & MXS & QC &  SEAM  & ZSS & Avg \\\hline\rowcolor{white} \multicolumn{8}{c}{Qwen2-1.5B-Instruct}\\\hline
GPT-3.5-Turbo  & 37.0&	33.2	&\textbf{94.4}	&	\textbf{83.3}&	10.0	&24.4&	42.3 \\
Llama3.1-8B+PPO & \textbf{38.5}	&\textbf{34.0}&	94.3	&	81.9	&\textbf{11.3}	&\textbf{24.8}	&\textbf{43.5} \\  \hline
  \rowcolor{white}\multicolumn{8}{c}{Qwen2-7B-Instruct}\\\hline
GPT-3.5-Turbo  & 44.1	&45.4&	94.7	&	82.7	&8.9	&25.9	&43.0\\
Llama3.1-8B+PPO & \textbf{46.7}&	\textbf{47.8}	&\textbf{95.1}	&	\textbf{82.5}	&\textbf{9.1}	&\textbf{26.9}	&\textbf{45.3} \\  \hline
  \rowcolor{white}\multicolumn{8}{c}{LLAMA3.1-8B-Instruct}\\\hline
GPT-3.5-Turbo  & 40.9	&\textbf{41.6}	&\textbf{95.8}	&	87.6	&\textbf{11.9}	&30.7&	50.3\\
Llama3.1-8B+PPO & \textbf{41.5}	&40.6&	94.9	&	\textbf{93.5}&	11.8	&\textbf{32.1}	&\textbf{53.1} \\  
\bottomrule
\end{tabular} 
\caption{\method performance when \methodrm is used to train a custom MD instruction generator model.}\label{tab:ppotable}
\vspace{-2mm}
\end{table}

\subsection{Impact on Advanced Model Capabilities}  \label{app:generalcapabilities}

We assess the impact of \method on general LLM capabilities including language understanding, advanced reasoning, mathematics, and coding by evaluating our models at the 7B and 8B parameter scales on GSM8K \citep{cobbe2021gsm8k}, MBPP \citep{austin2021programsynthesislargelanguage}, and MMLU \citep{hendrycks2021ethics, hendryckstest2021} via the OLMES framework \citep{gu2024olmesstandardlanguagemodel}. Results are shown in Table \ref{generaltab}. We observe that MDCure’d models achieve comparable average performance across the three benchmarks versus non-MDCure’d base models, demonstrating that \method data enhances MD modeling abilities of LLMs while preserving other advanced capabilities as well.

\section{Conclusion}
In this work, we presented \method, a framework for scalably and inexpensively generating high-quality MD post-training data to improve any base LLM’s ability to leverage long-range dependencies for MD tasks. \method addresses the challenge of data scarcity in MD task domains without depending on proprietary LLMs and is robust across model families and sizes. We further introduced \methodrm, a novel RM-as-a-Judge LLM to filter MD instruction data more cheaply and effectively than GPT-3.5-Turbo and enable moderately-sized LLMs to excel over SOTA proprietary counterparts as generators of high-quality MD instruction data. Our extensive experiments on various LLMs in many task and content domains show \method adds value beyond typical IT and substantially boosts MD performance while overcoming limitations of pre-training-based approaches. Future work could explore extending \method to generate instructions over additional topic domains.
\begin{table}[t]
\centering\footnotesize\setlength{\tabcolsep}{2.7pt}
\begin{tabular}{lcccc}
\toprule
Model	& GSM8K	& MBPP	& MMLU & Avg\\\midrule
Qwen2-7B-Instruct &	0.781	& 0.634	&	\textbf{0.706} & 0.707\\
+\method & \textbf{0.801}		& \textbf{0.668}	&	0.694 & \textbf{0.721}\\\midrule
LLAMA3.1-8B-Instruct	&0.800	&0.669	&	\textbf{0.717 }&0.729\\
+\method	&\textbf{0.806}	&\textbf{0.720}	&0.713	&\textbf{0.746}\\
\bottomrule
\end{tabular}
\caption{General performance of MDCure'd models.}\label{generaltab}
\vspace{-2mm}
\end{table}		
\section{Limitations} 

As abstractive tasks are the focus of many downstream MD applications, \method focuses on instruction generation for abstractive settings; future work can consider incorporation of extractive instruction templates to expand MDCure’s utility. Additionally, the design and application of our approach is limited to texts in English; future work can also consider exploring the utility of \method for non-English tasks. As observed with existing state-of-the-art generative language models, post-trained models may still suffer from factuality errors during generation. Although the focus of this work is not on factuality, we adopt measures to improve factual consistency by eliminating non-factual or unfaithful instruction generations in the Filtering stage; faithfulness aspects of our pipeline could be the subject of other future investigation. Lastly, potential improvements may arise from generating instructions and answers separately as opposed to together \citep{zhao2024selfguide}, from incorporation of short-context data during instruction-tuning \citep{bai2024longalign,gao2024trainlongcontextlanguagemodels}, and with inclusion of document sets in other text genres. Such design choices are left as possibilities for future research given the cost-effectiveness, efficacy, and domain-general nature of our approach.

\section{Ethics Statement}
As with any use of LLMs, there is a need to ensure the safety of our post-trained models’ responses through red-teaming or other safety measures. Use of our factuality criterion in assessing instruction data reduces the chances for \method to lead to unfactual examples due to hallucination during instruction generation. However, it is generally difficult to guarantee factual consistency during generative language modeling without further fine-tuning. Beyond this, we anticipate our work will improve LLMs’ efficacy in processing long documents and large document collections, which can accelerate value generation in domains such as business analytics and scientific literature mining.

\section*{Acknowledgments}
{This material is based upon work supported by the National Science Foundation Graduate Research Fellowship Program under Grant No. DGE-2139841.
We are grateful for the compute support provided by Google. Any opinions, findings, and conclusions or recommendations expressed in this material are those of the author(s) and do not necessarily reflect the views of the National Science Foundation or Google.}

\nocite{ernst-etal-2022-proposition,pmlr-v97-chu19b, cattan-etal-2021-realistic, eirew-etal-2021-wec,kwan-etal-2024-m4le, xu2024retrieval, wu2024longcontextalignmentshort, mao-etal-2020-multi, zhao2024longagentscalinglanguagemodels, agrawal2024cantrememberdetailslong, wang2024adalevalevaluatinglongcontextllms, an2024makellmfullyutilize, an2024trainingfreelongcontextscalinglarge, GHADIMI2023119308, chen-etal-2024-long, ge2024littlegoeslongway, yen2024helmetevaluatelongcontextlanguage, chan2024balancingcosteffectivenesssynthetic, kocisky2018narrativeqa,angelidis-etal-2021-extractive, dasigi2021qasper, huang2021govreport, zhong2021qmsum,chen-etal-2022-summscreen,kryscinski-etal-2022-booksum, pang-etal-2022-quality,shaham-etal-2022-scrolls, trivedi-etal-2022-music, wang-etal-2022-squality}
\bibliography{anthology,latex/custom,latex/no_cite}

\appendix

\section{Additional Ablation Results}  \label{app:ablations}

\subsection{Effect of Generation Approach} \label{analysis:prompteffect}
The efficacy of \method is dependent on the quality of initial candidate instruction generations. To this end, we conduct an ablation study to investigate the impact of several different prompting approaches on downstream instruction-tuned model performance. We consider only the zero-shot prompting setting given the high cost of prompting GPT models with few-shot examples that each contain multiple long documents. 

In our preliminary investigation, we find that prompt templates emphasizing the MD nature of generated instructions tend to yield higher-scoring results (e.g., suggesting penalties if generated instructions concern only a single document). Thereafter, we explore the utility of providing general vs. highly detailed specifications for the style of the generated instruction-answer pairs. We ultimately converge on two classes of prompt templates which we refer to as either \textit{General} or \textit{Style-Specific}. Details are given in in App. \ref{genprompts}. 

We examine the impact of General vs. Style-Specific prompt templates by fine-tuning FlanT5-Base and Qwen2-{7}B-Instruct using \methodrm-filtered datasets created using templates from one or both categories. As in \S \ref{sec:opensourcegenerator}, we limit the number of samples to 12K given the findings in \S \ref{analysis:data_scale}. When both templates are used, we employ a 1:3 ratio of General:Style-Specific instructions, found to be most effective in initial experiments. 

Results are shown in Table \ref{prompt_type}. We observe that use of General or Style-Specific templates alone to produce instructions does not always lead to downstream MD performance improvements (e.g., FlanT5-Base IT’ed with General-based instructions does not surpass base model performance on WikiHop). In comparison, combining the two gives the best performance gains in a consistent fashion across benchmarks.

\begin{table}
\centering\footnotesize\setlength{\tabcolsep}{3pt}
\renewcommand{\arraystretch}{1.2}
\begin{tabular}{@{}lcccccc@{}}\toprule
  & HQA & WH & MXS & QC &  SEAM & ZSS \\\hline
 \rowcolor{white} \multicolumn{7}{c}{FlanT5-Base}\\\hline
  Baseline (No IT) & 4.4 &45.1  & 38.7 & 48.0 & 1.7 & 13.1 \\  
General & 42.3 & 40.5 & 90.9 & 52.0 & 1.1 & 20.6 \\  
Style-Specific  & 37.9 & 48.1 & 91.8 & 48.9 & 1.3 & 20.0 \\  
Both   & \textbf{43.8 }& \textbf{48.4} & \textbf{92.4 }& \textbf{54.0} &  \textbf{1.9} & \textbf{22.0} \\\hline
 \rowcolor{white} \multicolumn{7}{c}{Qwen2-7B-Instruct}\\\hline
    Baseline (No IT) & 30.5 & 39.6 & 95.6 & 79.3 & 7.4 & 23.9 \\  
General & 40.8 & 44.9 & 93.8 & 81.3&  7.8 & 24.2 \\  
Style-Specific  & 40.2 & \textbf{45.8} & 94.2 & 80.7 & 7.8 & 23.7 \\  
Both   & \textbf{44.1} & 45.4 & \textbf{94.7} & \textbf{82.7}& \textbf{8.9}  &  \textbf{25.9}\\\bottomrule
\end{tabular}
\caption{Efficacy of different prompt approaches.} \label{prompt_type}
\end{table}

\subsection{Ablations on Scoring Criteria}  \label{ablationsection}

To justify our choice of scoring criteria in \methodrm, we conduct an ablation study to show the advantage of combining all six criteria, rather than using a subset of them. We also evidence the benefit of applying greater weight to MD-specific criteria when computing the overall score for each candidate instruction, as opposed to computing an evenly weighted sum (details in App. \ref{scoreweight}). {Ablation experiments were run using FlanT5-Base and Qwen2-7B-Instruct.}
Table \ref{ablation} shows the results. Across all benchmarks, filtering instruction data using a weighted average of six scoring criteria with emphasis on multi-document factors yields the best results. Further, removing any one criterion or utilizing even weighting across criteria leads to worsened IT’ed performance.

\begin{table}%
\centering 
\renewcommand{\arraystretch}{1.2}\scriptsize\setlength{\tabcolsep}{3.7pt}
\begin{tabular}{@{}lcccccc@{}}\toprule
  & HQA & WH & MXS & QC &  SEAM & ZSS \\\hline
   \rowcolor{white} \multicolumn{7}{c}{FlanT5-Base}\\\hline
\textit{all criteria, MD emphasis} & \textbf{43.8} & \textbf{48.4} & \textbf{92.4}	& 	\textbf{54.0} & \textbf{1.9}  & \textbf{22.0} \\
\textit{all criteria, evenly weighted} & 43.4 & 47.1 & 92.4	& 	52.2 &  1.8 & 21.2 \\
\textit{without Relevance} & 37.6 & 46.1 & 91.6	& 	50.2 &  1.6 & 20.4  \\
\textit{without Coh. \& Fac.} & 40.1 & 46.0 & 92.0	& 	52.2 &  1.5 & 20.4\\
\textit{without Creativity} & 38.4 & 44.2 & 92.2	& 	51.8 &  1.3 & 20.8\\
\textit{without Context Int.} & 14.1 &  45.4 & 85.3	& 	45.8 & 1.5  & 17.0 \\
\textit{without Inter-Doc. Rel.} & 40.9 & 45.8 & 91.1	& 	53.6 & 1.5  & 21.5 \\
\textit{without Complexity}  & 40.2 & 45.5 & 91.3	& 	54.2 &  1.4 & 20.0\\\hline
   \rowcolor{white} \multicolumn{7}{c}{Qwen2-7B-Instruct}\\\hline
   \textit{all criteria, MD emphasis} & \textbf{44.1} &\textbf{ 45.4} & 94.7	& 	\textbf{82.7} & \textbf{8.9} & \textbf{25.9} \\  
\textit{all criteria, evenly weighted} & 41.3 & 43.1 & 88.2	& 	68.9 &  8.5 & 25.4 \\
\textit{without Relevance} &37.5 & 40.7 & 94.4	& 	82.0 &  8.1 &  25.0 \\
\textit{without Coh. \& Fac.} & 41.9 & 43.8 & \textbf{94.9}	& 	80.7 &  8.4 & 23.2 \\
\textit{without Creativity} & 43.6 & 44.3 & 94.0	& 	\textbf{82.7} &  \textbf{8.9} & 25.4 \\
\textit{without Context Int.} & 37.1 & 42.4 & 94.2	& 	80.4 &  8.5 &  25.0\\
\textit{without Inter-Doc. Rel.} & 39.0 & 42.4 &94.2	& 	81.1 &  8.5 & 25.7 \\
\textit{without Complexity}  & 37.1 & 41.8 & 94.4	& 	81.1 &  8.5 & 25.7 \\\bottomrule
\end{tabular} 
\caption{Scoring criteria ablation study results.}\label{ablation}
\end{table}

\section{Dataset Construction}  \label{app:A}

\subsection{Source Texts}\label{newsheadappendix}
As noted in \S \ref{generationphase}, our experiments base data generation on the NewSHead corpus \citep{headline2020}. NewSHead consists of 369940 clusters of 3-5 related English news articles curated for the task of news story headline generation. We utilize the already-preprocessed version of the corpus supplied by \citet{caciularu-etal-2023-peek} at \url{https://storage.googleapis.com/primer_summ/processed_pre-training_data.tar.gz}. This follows the pre-processing procedure described in \citet{xiao-etal-2022-primera}. 

\subsection{Data Generation Prompts} \label{genprompts}
During the Generation phase of \method, we employ two types of prompt templates to encourage the generator model to produce a diverse set of instruction data. We distinguish these templates as either General or Style-Specific. 

General templates are inspired by \citet{koksal2023longform} and broadly ask the generator model to produce an instruction-answer pair adhering to a certain answer length (e.g., 1-2 words, 5 or more sentences). Emphasis is placed on creation of instructions that cannot be answered if any one context document is removed, and we employ various phrasings of this specification. 
We utilize a total of 14 general prompts. General templates A-D (Figure \ref{templates1}) are inspired by recent work that demonstrates the efficacy of grounding synthetic generations in pre-selected answer content from a given source. Modeling after \citep{koksal2023longform}, for these 4 prompt templates, we consider all possible combinations of two related documents in NewSHead and split the contents of every possible document pair into texts 1-3 sentences in length. Each segment is embedded via HuggingFace Sentence Transformers \citep{reimers-2019-sentence-bert} and aligned with the others via cosine similarity. Distinct segments are then paired according to highest alignment for according use in the prompt templates. To ensure segments are not too similar, we set a maximum similarity threshold of 0.70 per pair. General templates E-N (Figures \ref{template2} and \ref{templates3}) consider 2 or more context documents per generation.

On the other hand, Style-Specific templates (Figure \ref{ss1}) are adapted from \citet{zhu2024fannoaugmentinghighqualityinstruction} and employ a single phrasing while incorporating varying combinations of constraints regarding instruction complexity (e.g., analysis, recall), type (e.g., paraphrasing, inference), style (e.g., imperative, interrogative), and output length (e.g., 3-4 words, 3 sentences), with pre-defined options for each specification category. The aim is to enforce constraints on diversity of the generated candidate data. As defined below, we consider 4 characteristics, 4 types, and 3 styles of instruction, in combination with 8 possible answer lengths (Figure \ref{ss2}). Each possible resulting prompt template considers 2 or more context documents per generation.

To ensure a balanced composition for the pool of candidate instructions, we sample uniformly from all possible prompt templates during generation for each cluster of related documents. 

\subsection{Data Generation Hyperparameters}
To elicit high-quality candidate instructions in a cost-effective manner for our experiments, we utilize GPT-3.5-Turbo as the generator model in \method with all inference hyperparameters set to their default values. The cost is approximately \$25.00 per 35,000 candidate instruction generations. 

\subsection{Instruction Format} \label{app:instructionformat}

\begin{figure}[h!]
\centering
\includegraphics[width=\linewidth]{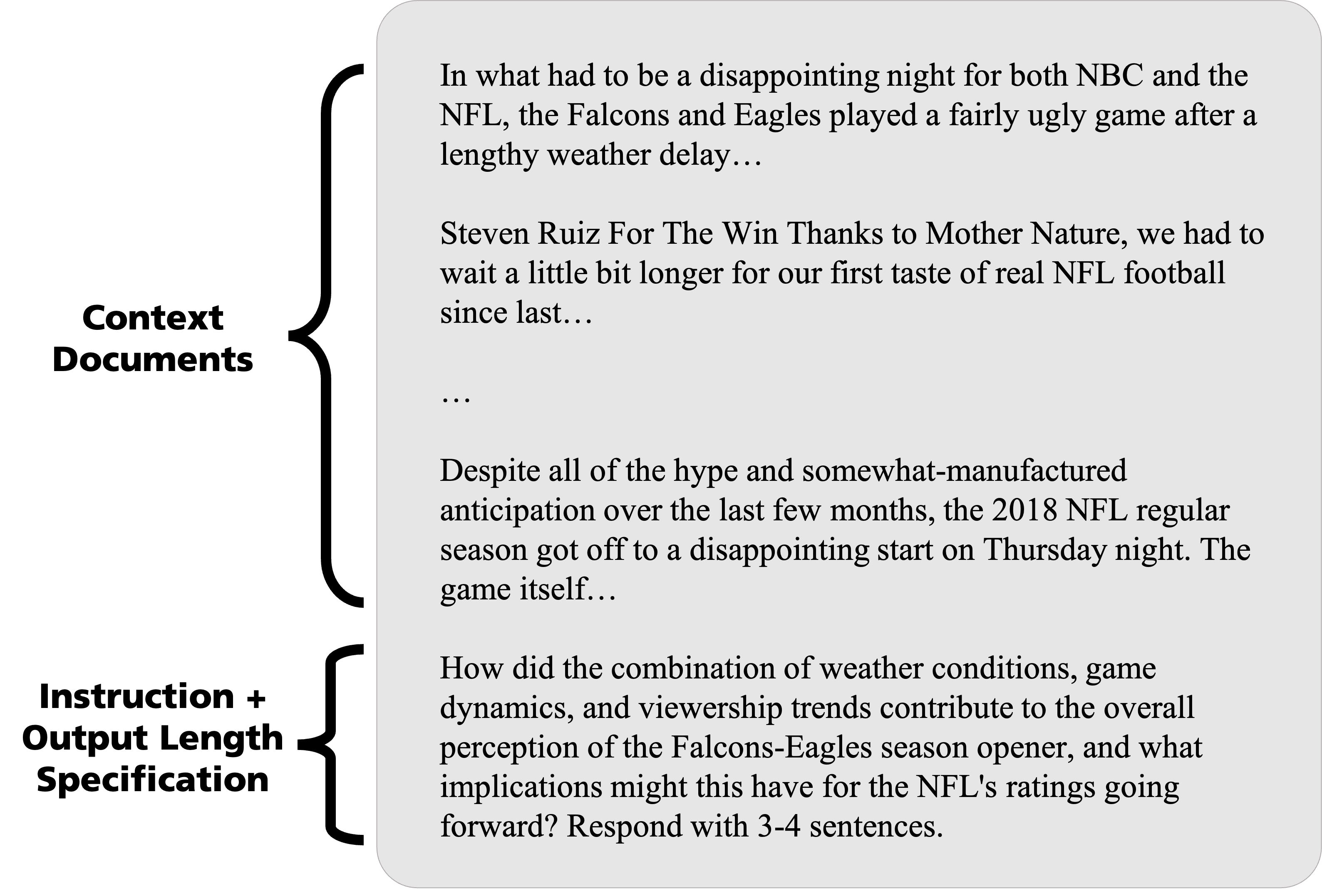}
\caption{Illustrative example of input format.} \label{fig:input-format}
\end{figure}

We format the MD instruction inputs to place the generated instruction after the source documents,
but alternate formats may be equally effective. 
Once all instruction-answer pairs are generated, since the expected output length of downstream MD tasks has a high variance (e.g., keyword based QA vs. long-form summarization), we append a brief direction indicating the expected length of the answer (i.e., expected number of words or sentences).\footnote{
Our preliminary experiments demonstrated non-negligible performance gains on downstream MD tasks with the attachment of such length direction. 
} 
Options for this direction differ depending on whether an instruction was created using a General or Style-Specific template (Figure \ref{ss3}). As shown in Fig. \ref{fig:input-format}, the direction is appended to the end of the finalized instruction input, separated by a space.

\begin{figure*}
\begin{tcolorbox}[colframe=black, colback=gray!5, boxrule=0.5pt, arc=2mm, width=\textwidth, left=1mm, right=1mm, top=1mm, bottom=1mm,title=General Template A]
Snippets: `\textcolor{blue}{\{segment1\}}', `\textcolor{blue}{\{segment2\}}'\\
Context Paragraphs: `\textcolor{blue}{\{doc1\}}', `\textcolor{blue}{\{doc2\}}'\\
Based on the given snippets and context paragraphs, construct an instruction-answer pair such that (1) the answer is based on the two snippets and (2) the instruction is a plausible prompt or question to which the answer would be the expected response. Make sure both snippets are required to answer the instruction. You will be penalized if the instruction concerns only one snippet. Format your response as:\\Instruction: <prompt or question>\\Answer: <answer>
\end{tcolorbox}

\begin{tcolorbox}[colframe=black, colback=gray!5, boxrule=0.5pt, arc=2mm, width=\textwidth, left=1mm, right=1mm, top=1mm, bottom=1mm,title=General Template B]
Snippets: `\textcolor{blue}{\{segment1\}}', `\textcolor{blue}{\{segment2\}}'\\
Based on the given snippets, construct an instruction-answer pair such that (1) the answer is yes and (2) the instruction is a plausible prompt or question to which yes would be the expected response. Make sure the answer does not conflict with the information in the snippets. You will be penalized if the instruction-answer pair is unfactual. Do NOT mention the snippets in the instruction. Format your response as:\\Instruction: <prompt or question>\\Answer: <yes>
\end{tcolorbox}
\begin{tcolorbox}[colframe=black, colback=gray!5, boxrule=0.5pt, arc=2mm, width=\textwidth, left=1mm, right=1mm, top=1mm, bottom=1mm,title=General Template C]
Snippets: `\textcolor{blue}{\{segment1\}}', `\textcolor{blue}{\{segment2\}}'\\
Based on the given snippets, construct an instruction-answer pair such that (1) the answer is no and (2) the instruction is a plausible prompt or question to which no would be the expected response. Make sure the answer does not conflict with the information in the snippets. You will be penalized if the instruction-answer pair is unfactual. Do NOT mention the snippets in the instruction. Format your response as:\\Instruction: <prompt or question>\\Answer: <no>
\end{tcolorbox}
\begin{tcolorbox}[colframe=black, colback=gray!5, boxrule=0.5pt, arc=2mm, width=\textwidth, left=1mm, right=1mm, top=1mm, bottom=1mm,title=General Template D]
Snippets: `\textcolor{blue}{\{segment1\}}', `\textcolor{blue}{\{segment2\}}'\\
Context Paragraphs: `\textcolor{blue}{\{doc1\}}', `\textcolor{blue}{\{doc2\}}'\\
Based on the given snippets and context paragraphs, construct an instruction-answer pair such that (1) the answer is a brief phrase and NOT a sentence and (2) the instruction is a plausible prompt or question to which the answer is the expected response. Make sure both snippets are required to answer the instruction. You will be penalized if the instruction concerns only one snippet. Make sure the answer is a brief phrase less than 7 words in length, with NO periods. You will be penalized if the answer is longer than 7 words or if the answer is a sentence. Format your response as:\\Instruction: <prompt or question>\\Answer: <answer>
\end{tcolorbox}
\caption{General Prompt Templates A-D}\label{templates1}
\end{figure*}
\begin{figure*}
\begin{tcolorbox}[colframe=black, colback=gray!5, boxrule=0.5pt, arc=2mm, width=\textwidth, left=1mm, right=1mm, top=1mm, bottom=1mm,title=General Template E]
Context Paragraphs: `\textcolor{blue}{\{doc1\}}', `\textcolor{blue}{\{doc2\}}'\\
Based on the two given context paragraphs, construct an instruction-answer pair such that (1) the answer is summary of the two paragraphs and (2) the instruction is a plausible prompt or question to which the answer is the expected response. Make sure both paragraphs are required to answer the instruction. You will be penalized if the instruction concerns only one paragraph. Make sure the answer does not conflict with the information in the paragraphs. You will be penalized if the instruction-answer pair is unfactual. Make sure the answer is at least 5 sentences in length. Do not mention the context paragraphs in the instruction. Format your response as:\\Instruction: <prompt or question>\\Answer: <answer>
\end{tcolorbox}
\begin{tcolorbox}[colframe=black, colback=gray!5, boxrule=0.5pt, arc=2mm, width=\textwidth, left=1mm, right=1mm, top=1mm, bottom=1mm,title=General Template F]
Context Paragraphs: `\textcolor{blue}{\{doc1\}}', `\textcolor{blue}{\{doc2\}}'\\
Based on the two given context paragraphs, construct an instruction-answer pair such that (1) the answer is summary of the two paragraphs and (2) the instruction is a plausible prompt or question to which the answer is the expected response. Make sure both paragraphs are required to answer the instruction. You will be penalized if the instruction concerns only one paragraph. Make sure the answer does not conflict with the information in the paragraphs. You will be penalized if the instruction-answer pair is unfactual. Make sure the answer is less than 5 sentences in length. Do not mention the context paragraphs in the instruction. Format your response as:\\Instruction: <prompt or question>\\Answer: <answer>
\end{tcolorbox}
\begin{tcolorbox}[colframe=black, colback=gray!5, boxrule=0.5pt, arc=2mm, width=\textwidth, left=1mm, right=1mm, top=1mm, bottom=1mm,title=General Template G]
\textcolor{blue}{\{context\_docs\}}\\A question or command that can ONLY be answered by utilizing ALL of the above documents and that CANNOT be answered if any one document is removed is:\\Question: <respond here>\\Answer: <respond here briefly>
\end{tcolorbox}
\begin{tcolorbox}[colframe=black, colback=gray!5, boxrule=0.5pt, arc=2mm, width=\textwidth, left=1mm, right=1mm, top=1mm, bottom=1mm,title=General Template H]
\textcolor{blue}{\{context\_docs\}}\\What is a question or command that can ONLY be answered by utilizing ALL of the above documents and that CANNOT be answered if any one document is removed?\\Question: <respond here>\\Answer: <respond here briefly>
\end{tcolorbox}
\begin{tcolorbox}[colframe=black, colback=gray!5, boxrule=0.5pt, arc=2mm, width=\textwidth, left=1mm, right=1mm, top=1mm, bottom=1mm,title=General Template I]
Articles:\\\textcolor{blue}{\{context\_docs\}}\\What is an exam question that can ONLY be answered by utilizing ALL of the above documents and that CANNOT be answered if any one document is removed?\\Exam Question: <respond here>\\Answer: <respond here briefly>
\end{tcolorbox}
\caption{General Prompt Templates E-I}\label{template2}
\end{figure*}
\begin{figure*}
\begin{tcolorbox}[colframe=black, colback=gray!5, boxrule=0.5pt, arc=2mm, width=\textwidth, left=1mm, right=1mm, top=1mm, bottom=1mm,title=General Template J]
\textcolor{blue}{\{context\_docs\}}\\What is a question or command that can ONLY be answered by utilizing ALL of the above documents and that CANNOT be answered if any one document is removed?\\Question: <respond here>\\Answer: <respond here, feel free to use a single word or phrase>
\end{tcolorbox}
\begin{tcolorbox}[colframe=black, colback=gray!5, boxrule=0.5pt, arc=2mm, width=\textwidth, left=1mm, right=1mm, top=1mm, bottom=1mm,title=General Template K]
\textcolor{blue}{\{context\_docs\}}\\A question or command that can ONLY be answered by utilizing ALL of the above documents and that CANNOT be answered if any one document is removed is:\\Question: <respond here>\\Answer: <respond here>
\end{tcolorbox}
\begin{tcolorbox}[colframe=black, colback=gray!5, boxrule=0.5pt, arc=2mm, width=\textwidth, left=1mm, right=1mm, top=1mm, bottom=1mm,title=General Template L]
\textcolor{blue}{\{context\_docs\}}\\What is a question or command that can ONLY be answered by utilizing ALL of the above documents and that CANNOT be answered if any one document is removed?\\Question: <respond here>\\Answer: <respond here, using ONLY a single word or phrase>
\end{tcolorbox}
\begin{tcolorbox}[colframe=black, colback=gray!5, boxrule=0.5pt, arc=2mm, width=\textwidth, left=1mm, right=1mm, top=1mm, bottom=1mm,title=General Template M]
Articles:\\\textcolor{blue}{\{context\_docs\}}\\Contrasting Question: <respond here>\\Answer: <respond here briefly>
\end{tcolorbox}
\begin{tcolorbox}[colframe=black, colback=gray!5, boxrule=0.5pt, arc=2mm, width=\textwidth, left=1mm, right=1mm, top=1mm, bottom=1mm,title=General Template N]
Articles:\\\textcolor{blue}{\{context\_docs\}}\\What is an exam question that can ONLY be answered by utilizing ALL of the above documents and that CANNOT be answered if any one document is removed?\\Exam Question: <respond here>\\Answer Choices: <respond here>\\Answer: <answer letter only>
\end{tcolorbox}
\caption{General Prompt Templates J-N}\label{templates3}
\end{figure*}

\begin{figure*}
\begin{tcolorbox}[colframe=black, colback=gray!5, boxrule=0.5pt, arc=2mm, width=\textwidth, left=1mm, right=1mm, top=1mm, bottom=1mm,title=Style-Specific Template]
You're proficient in crafting complex questions. Generate only one question and one answer that adheres to the provided \#Articles\#. Make sure the question and answer are factually consistent with the \#Articles\#. The question should meet the following criteria: \\
0. The person answering the question cannot see the \#Articles\#, so the question must not contain phrases like `Given the information provided', `Based on the provided information', or similar expressions that imply direct citations or references from \#Articles\#. \\
1. The question must REQUIRE synthesis of information in at least 2 of the provided documents in order to answer correctly. The more documents are involved the better. Ideally all documents are required to answer the question, such that losing any one of them will lead person answering the question to provide an incorrect response. You will lose your job if this criterion is not satisfied.\\
2. \textcolor{blue}{\{complexity\}}\\
3. \textcolor{blue}{\{type\}}\\
4. \textcolor{blue}{\{style\}}.\\
5. It requires \textcolor{blue}{\{answer\_length\}} to answer correctly.\\

The answer must be \textcolor{blue}{\{answer\_length\}} in length.\\

\#\#\# Articles: \\
\textcolor{blue}{\{context\_docs\}}\\

Question: <respond here>\\
Answer: <respond here>\\
\end{tcolorbox}
\caption{Style-Specific Prompt Template.}\label{ss1}
\end{figure*}
\begin{figure*}
\begin{tcolorbox}[colframe=black, colback=gray!5, boxrule=0.5pt, arc=2mm, width=\textwidth, left=1mm, right=1mm, top=1mm, bottom=1mm,title=Options for \{complexity\} Specification]
\begin{itemize}
    \item "It should be complex and require multiple-step reasoning across the documents to solve."
    \item "It demands critical thinking skills to analyze, evaluate, and synthesize multiple pieces of information from the different documents."
   \item  "It demands integrating knowledge from multiple documents to address its multifaceted nature."
   \item  "It should be simple and require only a few words to answer, yet utilize supporting evidence from at least 2 documents."
\end{itemize}
\end{tcolorbox}
\begin{tcolorbox}[colframe=black, colback=gray!5, boxrule=0.5pt, arc=2mm, width=\textwidth, left=1mm, right=1mm, top=1mm, bottom=1mm,title=Options for \{type\} Specification]
\begin{itemize}
    \item "It is a Natural language inference question: Assessing if evidence supports a conclusion."
      \item "It is a Paraphrasing question: Rewording a statement while retaining its meaning."
      \item "It is a Summarization question: Condensing key information from a larger text."
      \item "It is an Informational question: Locating a specific piece of information in the given evidence."
\end{itemize}
\end{tcolorbox}
\begin{tcolorbox}[colframe=black, colback=gray!5, boxrule=0.5pt, arc=2mm, width=\textwidth, left=1mm, right=1mm, top=1mm, bottom=1mm,title=Options for \{style\} Specification]
\begin{itemize}
    \item "It should be in the style of a command or imperative. For example, `Write a paragraph about...' or `Describe the...'"
      \item  "It should be in the style of a question or interrogative. For example, `What is the..?' or `How do you...?'"
      \item  "It should be in the style of a short phrase that serves as a query. For example, `today's forecast.' or `Donna’s car accident.'"
\end{itemize}
\end{tcolorbox}
\begin{tcolorbox}[colframe=black, colback=gray!5, boxrule=0.5pt, arc=2mm, width=\textwidth, left=1mm, right=1mm, top=1mm, bottom=1mm,title=Options for \{answer\_length\} Specification]
\begin{itemize}
    \item "1-2 words"
     \item "3-4 words"
     \item "a phrase of at least 5-6 words"
     \item "1-2 sentences"
     \item "3-4 sentences"
     \item "6 sentences"
     \item "8 sentences"
     \item "10 sentences"
\end{itemize}
\end{tcolorbox}
\caption{Detailed options for Style-Specific prompt template.}\label{ss2}
\end{figure*}

\begin{figure*}
\begin{tcolorbox}[colframe=black, colback=gray!5, boxrule=0.5pt, arc=2mm, width=\textwidth, left=1mm, right=1mm, top=1mm, bottom=1mm,title=Length Direction Options (General)]
\begin{itemize}
    \item "Answer briefly in 1-2 sentences."
     \item "Answer `yes' or `no'"
   \item "Answer `yes' or `no'"
      \item "Answer with a single word or brief phrase."
\item "Answer with at least 5 sentences."
\item "Answer with at most 5 sentences."
\end{itemize}
\end{tcolorbox}
\begin{tcolorbox}[colframe=black, colback=gray!5, boxrule=0.5pt, arc=2mm, width=\textwidth, left=1mm, right=1mm, top=1mm, bottom=1mm,title=Length Direction Options (Style-Specific)]
\begin{itemize}
    \item "Answer with \textcolor{blue}{\{answer\_length\}}."
   \item  "Answer using \textcolor{blue}{\{answer\_length\}}."
   \item  "Respond with \textcolor{blue}{\{answer\_length\}}."
   \item  "Respond using \textcolor{blue}{\{answer\_length\}}."
   \item  "Formulate your answer in \textcolor{blue}{\{answer\_length\}}."
   \item  "Reply with a \textcolor{blue}{\{answer\_length\}} answer."
   \item  "Craft your response in \textcolor{blue}{\{answer\_length\}}."
   \item  "Give a response that is \textcolor{blue}{\{answer\_length\}}."
   \item  "Answer in around \textcolor{blue}{\{answer\_length\}}."
\end{itemize}
\end{tcolorbox}
\caption{Options for direction regarding expected output length for instructions generated via General and Style-Specific templates.}\label{ss3}
\end{figure*}

\subsection{Sample Instruction Data}\label{app:sampledata}
We provide several qualitative examples demonstrating the diverse, complex, and open-ended nature of our \method data, formatted as prompt-response pairs (Table \ref{qualex}). Note that while each instruction input is formatted as discussed in App. \ref{app:instructionformat}, in Table \ref{qualex} we present only the generated prompt for each input and omit the context documents for easier visibility.

\begin{table}[h!]
\centering\scriptsize\setlength{\tabcolsep}{6pt}
\begin{tabular}{p{2.5cm}p{4.5cm}}
\toprule
\textbf{Instruction Prompt} & \textbf{Answer} \\\midrule
What is the standout feature of Samsung's Galaxy S10 fingerprint sensor identified by the majority of the articles? Answer with a single word or brief phrase. & Magical unlocking transition. \\\midrule
Describe the mechanism by which synthetic sandalwood stimulates hair growth. Give a response that is 3-4 words. & Activates receptor OR2AT4. \\\midrule
Describe the implications of NASCAR's new reduced horsepower package being used at the Auto Club 400 while considering the historical context of the race distance changes and past driver performances. Respond with 1-2 sentences. & The implementation of the reduced horsepower package at the Auto Club 400 marks a significant shift in NASCAR's strategy for intermediate tracks, aiming to enhance competitive balance; historically, the race has been shortened from 500 miles to 400 miles since 2011, impacting driver strategies and outcomes, as seen in past performances of drivers like Kevin Harvick and Kyle Busch.\\\midrule
Elaborate on the emotional and artistic significance of Adam Lambert's song "Feel Something" as it reflects his personal journey and professional struggles, highlighting how this piece serves as a precursor to his upcoming album. Craft your response in 6 sentences. & Adam Lambert's song "Feel Something" signifies a pivotal moment in his life, as it encapsulates his journey of overcoming personal struggles and creative disillusionment. Written during a phase of introspection, the track addresses Lambert's emotional experiences of feeling detached amidst the chaos of his career, illustrating his realization that the pursuit of love and connection could not be separated from the quest for personal happiness. The song represents both a cathartic release and an assertive step towards reclaiming his artistic vision, allowing Lambert to finally embrace the importance of self-care and vulnerability. This piece serves as the emotional cornerstone of his upcoming album, which aims to chronicle his commitment to taking ownership of his happiness and strength through music. The collaborative efforts with writers Benedict Cork and Josh Cumbee during its creation reflect a renewed artistic authenticity that Lambert seeks to reclaim. Thus, "Feel Something" not only marks a return to his musical roots but also signals a larger narrative of resilience, artistry, and self-discovery that will unfold in his forthcoming work. \\
\bottomrule
\end{tabular}
\caption{Representative instances of high-quality instruction-answer pairs, encompassing various MD task formulations and answer lengths, produced using our \method framework.}\label{qualex}
\end{table}	

\subsection{Diversity Analysis}\label{app:diversity}
To provide further insight regarding the syntactical and semantic diversity of our \method data, we follow the methodology used in \citet{wang-etal-2023-self-instruct, koksal2023longform} to analyze our size-12K \method dataset versus 12K samples from the training sets of HotpotQA-distractor and WikiHop, which are human-annotated MD benchmark datasets curated specifically for fine-tuning evaluation, along with 12K samples from the Tulu V2 SFT mixture \citep{rafailov2024directpreferenceoptimizationlanguage}, which is a mixture of human-annotated and synthetic data curated for general instruction tuning.

As in \citet{wang-etal-2023-self-instruct}, for each dataset, we parse the instruction in each training sample using Berkeley Neural Parser \citep{kitaev-klein-2018-constituency, kitaev-etal-2019-multilingual} and extract either the contained auxiliary and its dependent verb (e.g., ``do hear”) or the contained verb and its direct object (e.g., ``summarize events”). Additionally, as in \citet{zhu2024fannoaugmentinghighqualityinstruction}, we analyze the relative complexity of instruction-response pairs using the Deita-complexity-scorer model \citep{liu2023what}. We report statistics regarding unique verb-noun combinations and average complexity of each dataset in Table \ref{semstats}. 

\begin{table}[h!]
\centering\scriptsize\setlength{\tabcolsep}{6pt}
\begin{tabular}{p{1.7cm}p{1cm}p{1cm}p{1cm}p{1cm}}
\toprule
Dataset & \method & HotpotQA & WikiHop & Tulu V2\\\midrule
Unique Verbs & 210 & 302 & 248 & 197\\\midrule
Unique Noun+Verb/ Aux-Verb Pairs	& 1269 & 1392 & 552 & 422\\\midrule
Avg Complexity	& 2.65 & 2.07 & 1.60 & 2.10\\\midrule
Top 5 Verbs & describe, analyze, provide, summarize, reflect & direct, have, take, play, write & take, compose, use, get, make & answer, solve, ask, modify, explain\\
\bottomrule
\end{tabular}
\caption{Data statistics according to instruction composition.}\label{semstats}
\end{table}		

Versus human-annotated SFT and MD datasets, \method data covers a comparably wide range of tasks encompassing various instruction types, as indicated by its greater spread of unique verb+noun pairs, diverse and complex top 5 verbs, and highest-rated average complexity. These comparisons demonstrate the ability for \method prompts to yield generation of diverse MD instructions concerning tasks beyond those seen in typical SFT data, supporting the complementary nature of our instruction data and confirming the design of \method to reflect a broad array of task types and content.

\subsection{GPT Scoring Prompt}
We use the prompt shown in Figure \ref{gpt_prompt} to elicit effective scores from GPT-3.5-Turbo for the candidate generations in a zero-shot fashion. This prompt is adapted from that used in the self-curation step of \citet{li2024selfalignment}; other prompt variations (e.g., additive scoring as in \citep{yuan2024self}, rubric-free scoring as in \citep{li2023loogle}) yielded relatively worse MD performance in models instruction-tuned with the resulting filtered samples and were therefore discarded from consideration.
\begin{figure*}
\centering
    \begin{tcolorbox}[colframe=black, colback=gray!5, boxrule=0.5pt, arc=2mm, width=\textwidth, left=1mm, right=1mm, top=1mm, bottom=1mm]
Instruction:\\
\textcolor{blue}{\{generation\_prompt\}}\\

Response:\\
\textcolor{blue}{\{generated\_instruction\_answer\_pair\}}\\

Above are an Instruction from a user and a candidate Response from an AI Assistant. The goal of this AI Assistant is to generate a Response that effectively addresses the user's Instruction and that, in order to answer, requires the ability to reason over multiple documents. The Response should be a targeted question, instruction, prompt, or task that requires the use of information from different positions in the provided texts.\\

Evaluate whether the Response is a good example of how an AI Assistant should respond to the user's Instruction. Assign a score to the Response using the following 5-point scale:\\

1: It means the Response is incomplete, vague, off-topic, or not exactly what the user asked for in the Instruction. Perhaps it provides an incomplete prompt. Or it can be answered without looking at source documents provided in the Instruction. Or some content seems missing, the opening sentence repeats user's question, or it contains is irrelevant to the source documents or snippets provided in the Instruction.\\
2: It means the Response addresses some of the asks from the user but does not directly address the user's Instruction. For example, the Response only leverages one of several source documents or snippets provided in the Instruction. Or the Response can be answered using only ONE source document or snippet, and thus does not effectively assess and require use of multi-document reasoning capabilities.\\
3: It means the Response is fair and addresses all the basic asks from the user. It is complete and self contained and is relevant to most of the documents or snippets provided, but not all. It may be somewhat helpful toward assessing an agent's multi-document reasoning capability but still has room for improvement.\\
4: It means the Response is good quality. Specifically, the Response can only be answered by performing reasoning across most of the documents or snippets provided in the Instruction. The provided documents or snippets include all the information required to answer the Response, i.e. no information beyond that provided in the Instruction is needed. The Response has minor room for improvement, e.g. more concise and focused.\\
5: It means the Response is perfect, i.e. it can only be answered with strong ability to extract and synthesize information across the documents or snippets provided in the Instruction. The Response utilizes ALL documents or snippets provided in the instruction. The provided documents or snippets include all the information required to answer the Response. It is well-written and effective toward the AI Assistant's goal and has no irrelevant content.\\

Assess the Response and assign a rating score using this scale. Respond with "Score: <rating>".
\end{tcolorbox}
\caption{Prompt used for GPT-3.5-Turbo to evaluate the quality of candidate MD instruction-answer pairs.}
\label{gpt_prompt}
\end{figure*}

\subsection{\methodrm Training Data Prompt}\label{rmdataprompt}
As described in \S \ref{filtering}, we use the prompt shown in Figure \ref{pref_prompt} to generate fine-grained reward training data for \methodrm.
\begin{figure*}
   \centering 
\begin{tcolorbox}[colframe=black, colback=gray!5, boxrule=0.5pt, arc=2mm, width=\textwidth, left=1mm, right=1mm, top=1mm, bottom=1mm]
Instruction Quality Rating Task\\
Rate the quality of the generated instruction based on the provided documents, using a scale of 1-5. Only provide numerical scores, without any rationale or explanation.\\

Relevance: Does the instruction align well with the content of the documents? Does it make sense given the provided information?\\
Coherence \& Factuality: Is the instruction-answer pair coherent, logical, and factually accurate? Does the answer appropriately address the instruction and is it well-supported by the documents?\\
Creativity: How diverse or creative is the instruction in terms of question type (e.g., factual, inferential) and format (e.g., multiple-choice, open-ended)?\\
Context Integration: How well does the instruction leverage and synthesize information from multiple documents to form a comprehensive response?\\
Inter-Document Relationships: Does the instruction encourage understanding relationships (e.g., comparisons, contrasts, discrepancies) between different documents?\\
Complexity: Does the instruction appropriately challenge the answerer to think critically and synthesize information from multiple sources?\\

Input:\\
Context: \textcolor{blue}{\{context\_docs\}}\\
\textcolor{blue}{\{instruction\_sample\}}\\

Output (provide only numerical scores, no rationale): \\
Relevance: [score]\\
Coherence \& Factuality: [score]\\
Creativity: [score]\\
Context Integration: [score]\\
Inter-Document Relationships: [score]\\
Complexity: [score]\\
\end{tcolorbox}
\caption{Prompt to elicit fine-grained score training data for \methodrm.} \label{pref_prompt}
\end{figure*}

\subsection{\method Data Scoring}\label{scoreweight}
Given a candidate instruction-answer pair, \methodrm generates a vector of six floating point values in the range $[0,1]$. To re-scale each value to the 1-5 range, we multiply each criterion score by 4 and add 1. To obtain the overall score for the instruction sample, we take a weighted sum of the resulting values, using a weight of $\frac{1}{9}$ for the general quality criteria (Relevance, Coherence \& Factuality, Creativity) and a weight of $\frac{2}{9}$ for the multi-document specific criteria (Context Integration, Inter-Document Relationships, Complexity). In \S \ref{ablationsection}, we compare this weighting assignment with the evenly weighted alternative, which utilizes a weight of $\frac{1}{6}$ for each criterion in the case where all 6 criteria are used, and a weight of $\frac{1}{5}$ for each criterion in the case where one criterion is ablated at a time.

\subsection{\method Dataset Statistics} \label{datadeets}
We summarize statistics for the unfiltered, GPT-3.5-Turbo-filtered, and \methodrm-Filtered instruction data as shown in Table \ref{dataset_stats}. Quantitatively, we observe that the \methodrm-Filtered data tends to have a slightly higher number of documents per sample compared to the GPT-filtered data. Meanwhile, the instruction lengths in the \methodrm-Filtered and unfiltered pools are quite similar, both having longer and more detailed instructions on average, whereas the GPT-filtered pool features shorter instruction lengths. Complexity in terms of context length and instruction lengths is apparent across all data settings, but the \methodrm-Filtered data handles the most complex tasks while GPT-filtered data leans towards handling smaller, more straightforward tasks.
\begin{table}[h!]
\centering\scriptsize\setlength{\tabcolsep}{6pt}
\begin{tabular}{lccc}
\hline
Data Setting & \# Instr. & Avg. \# Context Docs & Avg. Instr. Length  \\
\hline
Unfiltered & 221,835 & 3.4  & 1,754  \\  
GPT-Filtered & 70,598 & 3.2 & 1,377 \\  
RM-Filtered  & 125,994 & 3.4 & 1,719 \\
\hline
\end{tabular}
\caption{Data statistics according to instruction composition.}\label{dataset_stats}
\end{table}

\subsection{Long-Context \& Few-Shot Data Construction} \label{longcontextdata}
For long-context and few-shot post-training experiments (\S \ref{sec_fs_32k}), we construct instruction inputs by incorporating distractor articles or few-shot examples, respectively. Distractor articles are identified by embedding all NewSHead documents using the Sentence Transformers model \texttt{all-distilroberta-v1},\footnote{We choose this embedding model from \url{https://www.sbert.net/docs/sentence_transformer/pretrained_models.html} due to its strong performance, long context length, and compatibility with cosine similarity.} and taking the most cosine-similar documents to the documents in the source cluster. We choose similar instead of random documents to increase task difficulty and encourage identification of relevant documents in a long context, as recent work finds using packing related texts during training can assist models to read and reason across text boundaries \citep{shi2023context}. For few-shot training, exemplars are chosen randomly, with each exemplar appearing once across the duration of training.

\section{Training Details} 

\subsection{Instruction Tuning}\label{trainingdeets}
All FlanT5 models are trained using 8 NVIDIA A6000 GPUs or 8 A100 40GB GPUs. For LLAMA3.1 and Qwen2 fine-tuning experiments, we employ the widely-used Unsloth repository with HuggingFace to perform 4-bit quantization and Low-Rank Adaptation (LoRA). We choose to use LoRA over full fine-tuning due to compute limitations and since preliminary experiments yielded similar performance in both settings. LoRA adapters are used for every linear layer for all query-, key-, value-, output-, gate-, up-, and down-projection matrices with a rank of 16, alpha parameter of 32, and zero adapter dropout. As Unsloth only accommodates single-GPU training, all LLAMA3.1 and Qwen2 models are trained on a single A6000 or A100 GPU. All fine-tuning experiments utilized gradient checkpointing aside from those involving FlanT5-Base.

As the FlanT5 models could be trained with a maximum length of 4096 tokens without GPU memory overflow, we set the maximum input length of the training data to 4096 for all fine-tuning experiments beyond those in \S \ref{sec_fs_32k} to ensure comparability. Any samples exceeding this length were truncated from the right, by shortening component context documents to equal lengths and preserving the entirety of the subsequent instruction. For fine-tuning experiments with long-context and few-shot MDCure data (\S \ref{sec_fs_32k}), we set the maximum input length to 32000 tokens.

All fine-tuning experiments utilized a global batch size of 64, AdamW optimization with $\beta_1=0.9$ and $\beta_2=0.98$, a linear learning rate schedule, warmup ratio of 0.15, dropout of 0.05, weight decay of 0.01, and the best of either 1e-4 or 5e-5 for the maximum learning rate. Fine-tuning is performed for 1100, 1650, and 2250 steps for the 12K-, 36K-, and 72K-training sample settings. For the long-context and few-shot training settings, we use a total of 36K samples, with the first 90\% of training steps utilizing 4096 maximum input length and the remaining 10\% utilizing 32000 maximum input length, in a similar fashion to \citet{dubey2024llama3herdmodels}. In terms of training data composition, unless otherwise stated, all models employed a 1:3 ratio of General:Style-Specific prompt template generations. Using the hardware specified above, all fine-tuning experiments require 24-120 GPU-hours.

In all settings, we determine the best checkpoint for each model by comparing checkpoint performances in the zero-shot prompting setting on 500 test samples from MultiNews, Multi-XScience, HotpotQA-distractor, WikiHop, and QMDSCNN; we use LLM evaluation to assess MDS performance and F1 and EM scores to assess MHQA performance. Evaluation details are provided in App. \ref{llmevaldeets}. For language modeling settings, we apply the loss to the instruction input as well during training, per recommendation by \citet{shi2024instruction}.

\subsection{MDCureRM Training} \label{rmdesign}

To train MDCure RM, we utilize 8 A100 40GB GPUs. Training occurs for 4 epochs using a training set of size approximately 17K samples and validation and test sets of size approximately 1.5K. We set the global batch size to 64 and use AdamW 8-bit optimization, a linear learning rate schedule, warmup ratio of 0.1. We perform a small hyperparameter search over learning rates of 7e-6, 1e-5, 5e-5, 1e-4, 7e-4, and 1e-3.

Beyond use of a single regression head to construct \methodrm as in \citet{armorm}, we investigate the impact of several variations in the architecture and design of MDCureRM. These variations include the addition of 2 or 3 linear layers to the backbone LLM as opposed to a single regression head, as well as use of an MLP head with attention pooling. Due to compute limitations we were unable to investigate the impact of unfreezing the backbone LLAMA3 model during \methodrm training. Across these variations, we determined the best scoring model based on test loss, finding use of a single regression head and learning rate 1e-3 most effective. Small-scale experiments on FlanT5-Base demonstrated this improves over the MLP variant.

\subsection{PPO Training} \label{app:ppo}

For PPO training of LLAMA3.1-8B-Instruct (\S\ref{sec:ppo}), we utilized 4 A100 80GB GPUs. Training occurred over 72K samples for a single epoch, with a global batch size of 64, learning rate of 1e-5, and LORA as described in \S\ref{trainingdeets}. Additional hyperparameter search was performed over learning rates of 1e-4 and 1e-3. All other hyperparameter settings remained consistent with default values.

\section{Evaluation Details}

\subsection{Benchmark Descriptions} \label{benchmarkdeets}
We provide details on the datasets used to benchmark our instruction-tuned models versus baselines. All benchmarks are in English.

HotpotQA \citep{yang-etal-2018-hotpotqa} is a multi-hop QA dataset consisting of approximately 113,000 questions and paragraphs sourced from Wikipedia, where for each sample there are eight paragraphs that serve as distractors and two paragraphs containing the information necessary to answer the given question. The objective is to identify the correct answer and evidence spans in the text. 

WikiHop \citep{welbl2018constructing} is a multi-hop QA dataset consisting of approximately 43,000 samples constructed from entities and relations from WikiData and supporting documents from WikiReading. Each sample provides a question, 2-79 potential answers, and 3-63 supporting contexts. The aim is to arrive at the correct answer based on the context and question.

Multi-XScience \citep{lu-etal-2020-multi-xscience} is a MDS dataset sourced from Arxiv and Microsoft scientific graphs. The aim is to produce the related works section for a query publication given the abstract of the query paper and several abstracts of other referenced works. Multi-XScience is considered relatively more difficult versus datasets such as Multi-News since it is less susceptible to positional and extractive biases.

QMDSCNN \citep{Pasunuru2021DataAF} is a query-focused MDS dataset of approximately 287K samples created by extending the CNN/DailyMail single-document summarization dataset to the multi-document setting. Given paragraph-scale chunks of various news articles, the aim is to return relevant context portions given queried article titles.

SEAM \citep{lior2024seamstochasticbenchmarkmultidocument} is a recent multi-document benchmark that addresses the sensitivity of LMs to prompt variations through repeated evaluations with controlled input-output formats on a diverse set of multi-document datasets. Tasks include multi-document summarization, multi-hop question-answering, and cross-document coreference resolution.

ZeroSCROLLS \citep{zeroscrolls} is a collection of datasets for long-context natural language tasks across domains of summarization, QA, sentiment classification, and information reordering. Successful performance on the benchmark requires synthesizing information across long texts. 

\subsubsection{Dataset Abbreviations} \label{abbreviations}
We provide a list of dataset name abbreviations in Table \ref{abbs}.
\begin{table}[h!]
    \centering\scriptsize\setlength{\tabcolsep}{6pt}
\begin{tabular}{ll}
\mediumhline
Dataset Name & Abbreviation \\
\hline
HotpotQA-distractor & HQA \\
WikiHop & WH \\
Multi-XScience & MXS \\
QMDSCNN & QC \\\hline
\rowcolor{Gray}\multicolumn{2}{c}{SEAM}\\\hline
MultiNews & MN \\
OpenAsp & OA \\
MuSiQue & MSQ \\
SciCo  & SC \\
ECB+ & ECB+ \\\hline
\rowcolor{Gray}\multicolumn{2}{c}{ZeroScrolls}\\\hline
GovReport & GvRp \\
SummScreenFD & SSFD \\
QMSum & QMsm \\
NarrativeQA &  NrQA\\
Qasper &  Qspr\\
QuALITY &  QLTY\\
SQuALITY & SQLT \\
MuSiQue &  MuSQ\\
SpaceDigest & SpDg \\
BookSumSort & BkSS \\
ZeroScrolls (ZS) Score & ZSS\\
\mediumhline
\end{tabular}
\caption{Dataset name abbreviations used for results tables in the main text.}\label{abbs}
\end{table}

\subsection{Baselines} \label{baselinedeets}

PRIMERA \citep{xiao-etal-2022-primera} is a 447M-parameter model initialized from LED-large \citep{beltagy2020longformerlongdocumenttransformer} and trained using an entity salience-based pre-training objective over the NewSHead dataset. It previously stood as the state-of-the-art in multi-document summarization. 

\textsc{QAMDen} \citep{caciularu-etal-2023-peek} is a 447-parameter model initialized in PRIMERA sparse attention style over LED-large and trained using a QA-based pre-training objective over the NewSHead dataset. It surpasses PRIMERA in performance on MDS, QMDS, and MDQA tasks.

LongAlign \citep{bai2024longalign} is a recipe for long-context alignment that enables improved long-context performance of fine-tuned models through a combination of data and packing strategies during training. It provides a series of instruction-tuned long-context models up to 13B parameters in size adapted from the ChatGLM3 \citep{glm2024chatglm} and LLAMA2 \citep{touvron2023llama2openfoundation} model families. We use LongAlign-7B for comparability to Qwen2-7B-Instruct and LLAMA3.1-8B-Instruct, which we use for \method experiments.

{ProLong-8B-64k-Instruct \citep{gao2024trainlongcontextlanguagemodels} is continued-trained and supervised-fine-tuned from the LLAMA-3-8B model, with a maximum context window of 64K tokens. It achieves state-of-the-art long-context performance versus similarly sized models and outperforms LLAMA3.1-8B-Instruct on many long-context tasks.}

Jamba 1.5 Mini \citep{jamba} is a speed-optimized model with 52B total parameters and 12B active parameters that is specifically trained for improved long-context handling and quality. It utilizes a hybrid transformer architecture.

\subsection{Zero-Shot Evaluation Prompts} \label{evaldeets}
To perform zero-shot prompting evaluations on Multi-XScience, QMDSCNN, HotpotQA, and WikiHop, we devise effective zero-shot prompts for each dataset in a similar fashion to \citet{zeroscrolls} and utilize the same input format as \citet{zeroscrolls} as shown in Figure \ref{zeroshotformat}. For each model family (FlanT5, LLAMA3.1, Qwen2), we choose the best one of 5 tailored prompt variations based on base model performance to utilize for evaluations of all subsequent instruction-tuned models. We display the final prompts utilized for each model family in Figure \ref{modelprompts}. During inference, we use greedy decoding and limit the model generation length to at most the 90th percentile of the maximum target output length. The input context is truncated to length 8192 for every model, and for all results reported in the main text, we utilize 1500 test samples per dataset as opposed to using the entire test set for each dataset to accommodate cost limitations imposed by LLM evaluations for MDS. 
\begin{figure*}
\centering
    \begin{tcolorbox}[colframe=black, colback=gray!5, boxrule=0.5pt, arc=2mm, width=\textwidth, left=1mm, right=1mm, top=1mm, bottom=1mm, title=Zero-Shot Input Format for HotpotQA]
\textcolor{blue}{\{prompt\}}\\

Question:\\
\textcolor{blue}{\{question\}}\\

Supporting Documents:\\
\textcolor{blue}{\{supporting\_documents\}}\\

Answer:
\end{tcolorbox}
    \begin{tcolorbox}[colframe=black, colback=gray!5, boxrule=0.5pt, arc=2mm, width=\textwidth, left=1mm, right=1mm, top=1mm, bottom=1mm, title=Zero-Shot Input Format for WikiHop]
\textcolor{blue}{\{prompt\}}\\

Documents:\\
\textcolor{blue}{\{supporting\_documents\}}\\

Question:\\
\textcolor{blue}{\{question\}}\\

Answer Candidates:\\
\textcolor{blue}{\{answer\_choices\}}\\

Answer:
\end{tcolorbox}
    \begin{tcolorbox}[colframe=black, colback=gray!5, boxrule=0.5pt, arc=2mm, width=\textwidth, left=1mm, right=1mm, top=1mm, bottom=1mm, title=Zero-Shot Input Format for Multi-XScience]
\textcolor{blue}{\{prompt\}}\\

Documents:\\
\textcolor{blue}{\{source\_and\_reference\_abstracts\}}\\

Related Work Section:
\end{tcolorbox}
    \begin{tcolorbox}[colframe=black, colback=gray!5, boxrule=0.5pt, arc=2mm, width=\textwidth, left=1mm, right=1mm, top=1mm, bottom=1mm, title=Zero-Shot Input Format for QMDSCNN]
\textcolor{blue}{\{prompt\}}\\

Query:\\
\textcolor{blue}{\{question\}}\\

Articles:\\
\textcolor{blue}{\{articles\}}\\

Summary:
\end{tcolorbox}
\caption{Task input formats used for zero-shot prompting evaluation on HotpotQA, WikiHop, Multi-XScience, and QMDSCNN, adapted from \citet{zeroscrolls}.}
\label{zeroshotformat}
\end{figure*}
\begin{figure*}
\centering
\begin{tcolorbox}[colframe=black, colback=gray!5, boxrule=0.5pt, arc=2mm, width=\textwidth, left=1mm, right=1mm, top=1mm, bottom=1mm, title=Final Task Prompts for FlanT5 Models]
\begin{itemize}
    \item HotpotQA: "You are given a question and multiple supporting documents separated by `|||||'. Answer the question as concisely as you can, using a single phrase if possible."
\item     WikiHop: "You are given multiple supporting documents separated by `|||||', a question, and list of answer candidates. Answer the question by selecting one of the provided answer candidates."
   \item  Multi-XScience: "You are given several scientific documents, separated by `|||||'. Write the related-work section of a paper based on the given abstracts and articles. Answer in a single paragraph."
   \item  QMDSCNN: "You are given a query and an article. Answer the query as concisely as you can by summarizing the relevant information from the article. Respond in a single paragraph."
\end{itemize}
\end{tcolorbox}
\begin{tcolorbox}[colframe=black, colback=gray!5, boxrule=0.5pt, arc=2mm, width=\textwidth, left=1mm, right=1mm, top=1mm, bottom=1mm, title=Final Task Prompts for LLAMA3.1 Models]
\begin{itemize}
    \item HotpotQA: "You are given a question and several supporting documents. Answer the question as concisely as you can, using a single word or phrase."
\item     WikiHop: "You are given multiple supporting documents separated by `|||||', a question, and list of answer candidates. Answer the question by selecting one of the provided answer candidates."
   \item  Multi-XScience: "You are given several scientific abstracts. Write a related works section based on the abstracts. Answer in a short paragraph. Be very brief."
   \item  QMDSCNN: "You are given a query and several articles. Answer the query as concisely as you can by supplying the relevant information from the articles. Respond briefly, in less than one paragraph."
\end{itemize}
\end{tcolorbox}
\begin{tcolorbox}[colframe=black, colback=gray!5, boxrule=0.5pt, arc=2mm, width=\textwidth, left=1mm, right=1mm, top=1mm, bottom=1mm, title=Final Task Prompts for Qwen2 Models]
\begin{itemize}
    \item HotpotQA: "You are given a question and multiple supporting documents separated by `|||||'. Answer the question as concisely as you can, using a single word or phrase."
\item     WikiHop: "You are given multiple supporting documents separated by `|||||', a question, and list of answer candidates. Answer the question by selecting one of the provided answer candidates."
   \item  Multi-XScience: "You are given several scientific abstracts, separated by `|||||'. Write the related works section of a paper based on the given abstracts. Answer in a single paragraph."
   \item  QMDSCNN: "You are given a query and several articles. Answer the query as concisely as you can by supplying the relevant information from the articles. Respond briefly, in less than one paragraph."
\end{itemize}
\end{tcolorbox}
\caption{Task prompts yielding the best performance for base models in the FlanT5, LLAMA3.1, and Qwen2 model families, selected from among 5 tailored variations.}
\label{modelprompts}
\end{figure*}

\subsection{LLM Evaluation Details} \label{llmevaldeets}

As noted in \S \ref{expsetup}, to evaluate MDS performance, we turn to LLM evaluation to overcome limitations of $n$-gram-based methods such as ROUGE in long-context settings \citep{zeroscrolls}. We employ the open-source LLM evaluation framework G-Eval \citep{liu-etal-2023-g}, which is one of the most highly human-correlated evaluation frameworks to date when paired with GPT-4. For a given document and summary, G-Eval determines the quality of the summary by issuing four fine-grained scores for Relevance, Coherence, Consistency, and Fluency. The overall LLM score for a given summarization response is obtained by taking the average of the four scores and normalizing the result to the 0-100 range. We adapt G-Eval to the multi-document setting using the prompts shown in Figures \ref{relevance}, \ref{coherence}, \ref{consistency}, and \ref{fluency}. 

\begin{figure*}
\centering
 \begin{tcolorbox}[colframe=black, colback=gray!5, boxrule=0.5pt, arc=2mm, width=\textwidth, left=1mm, right=1mm, top=1mm, bottom=1mm, title=G-Eval Prompt Template for LLM Evaluation 
 of MDS Relevance]

You will be given one summary written for a set of related documents.\\

Your task is to rate the summary on one metric.\\

Please make sure you read and understand these instructions carefully. Please keep this document open while reviewing, and refer to it as needed.\\

Evaluation Criteria:\\

Relevance (1-5) - selection of important content from the source. The summary should include only important information from the source documents. Annotators were instructed to penalize summaries which contained redundancies and excess information.\\

Evaluation Steps:\\

1. Read the summary and the source documents carefully.\\
2. Compare the summary to the source documents and identify the main points of the articles.\\
3. Assess how well the summary covers the main points of the source documents, and how much irrelevant or redundant information it contains.\\
4. Assign a relevance score from 1 to 5.\\

Example:\\

Source Text:\\

\textcolor{blue}{\{documents\}}\\

Summary:\\

\textcolor{blue}{\{summary\}}\\

Evaluation Form (scores ONLY):\\

- \{Relevance\}\\
\end{tcolorbox}
\caption{G-Eval prompt for NLG evaluation of the relevance of a given summarization response.}
\label{relevance}
\end{figure*}

\begin{figure*}
\centering
 \begin{tcolorbox}[colframe=black, colback=gray!5, boxrule=0.5pt, arc=2mm, width=\textwidth, left=1mm, right=1mm, top=1mm, bottom=1mm, title=G-Eval Prompt Template for LLM Evaluation 
 of MDS Coherence]

You will be given one summary written for a set of related documents.\\

Your task is to rate the summary on one metric.\\

Please make sure you read and understand these instructions carefully. Please keep this document open while reviewing, and refer to it as needed.\\

Evaluation Criteria:\\

Coherence (1-5) - the collective quality of all sentences. We align this dimension with the DUC quality question of structure and coherence whereby "the summary should be well-structured and well-organized. The summary should not just be a heap of related information, but should build from sentence to a coherent body of information about a topic.\\

Evaluation Steps:\\

1. Read the source documents carefully and identify the main topic and key points.\\
2. Read the summary and compare it to the source documents. Check if the summary covers the main topic and key points of the source documents, and if it presents them in a clear and logical order.\\
3. Assign a score for coherence on a scale of 1 to 5, where 1 is the lowest and 5 is the highest based on the Evaluation Criteria.\\

Example:\\

Source Text:\\

\textcolor{blue}{\{documents\}}\\

Summary:\\

\textcolor{blue}{\{summary\}}\\

Evaluation Form (scores ONLY):\\

- \{Coherence\}\\
\end{tcolorbox}
\caption{G-Eval prompt for NLG evaluation of the coherence of a given summarization response.}
\label{coherence}
\end{figure*}

\begin{figure*}
\centering
 \begin{tcolorbox}[colframe=black, colback=gray!5, boxrule=0.5pt, arc=2mm, width=\textwidth, left=1mm, right=1mm, top=1mm, bottom=1mm, title=G-Eval Prompt Template for LLM Evaluation 
 of MDS Consistency]

You will be given one summary written for a set of related documents.\\

Your task is to rate the summary on one metric.\\

Please make sure you read and understand these instructions carefully. Please keep this document open while reviewing, and refer to it as needed.\\

Evaluation Criteria:\\

Consistency (1-5) - the factual alignment between the summary and the summarized sources. A factually consistent summary contains only statements that are entailed by the source documents. Annotators were also asked to penalize summaries that contained hallucinated facts.\\

Evaluation Steps:\\

1. Read the source documents carefully and identify the main facts and details they present.\\
2. Read the summary and compare it to the source documents. Check if the summary contains any factual errors that are not supported by the source documents.\\
3. Assign a score for consistency based on the Evaluation Criteria.\\

Example:\\

Source Text:\\

\textcolor{blue}{\{documents\}}\\

Summary:\\

\textcolor{blue}{\{summary\}}\\

Evaluation Form (scores ONLY):\\

- \{Consistency\}\\
\end{tcolorbox}
\caption{G-Eval prompt for NLG evaluation of the consistency of a given summarization response.}
\label{consistency}
\end{figure*}

\begin{figure*}
\centering
 \begin{tcolorbox}[colframe=black, colback=gray!5, boxrule=0.5pt, arc=2mm, width=\textwidth, left=1mm, right=1mm, top=1mm, bottom=1mm, title=G-Eval Prompt Template for LLM Evaluation 
 of MDS Fluency]

You will be given one summary written for a set of related documents.\\

Your task is to rate the summary on one metric.\\

Please make sure you read and understand these instructions carefully. Please keep this document open while reviewing, and refer to it as needed.\\

Evaluation Criteria:\\

Fluency (1-3): the quality of the summary in terms of grammar, spelling, punctuation, word choice, and sentence structure.\\

- 1: Poor. The summary has many errors that make it hard to understand or sound unnatural.\\
- 2: Fair. The summary has some errors that affect the clarity or smoothness of the text, but the main points are still comprehensible.\\
- 3: Good. The summary has few or no errors and is easy to read and follow.\\

Evaluation Steps:\\

1. Read the summary.\\
2. Assign a score for fluency based on the Evaluation Criteria.\\

Example:\\

Source Text:\\

\textcolor{blue}{\{documents\}}\\

Summary:\\

\textcolor{blue}{\{summary\}}\\

Evaluation Form (scores ONLY):\\

- \{Fluency\}\\
\end{tcolorbox}
\caption{G-Eval prompt for NLG evaluation of the fluency of a given summarization response.}
\label{fluency}
\end{figure*}

As GPT4 is expensive, we investigate use of several other commercial LLMs with the adapted G-Eval framework for evaluation on Multi-XScience and QMDSCNN, with the goal of ensuring strong alignment of resulting ratings with human evaluation. To validate our choice of LLMs to evaluate MDS performance, we employ two expert annotators similarly to the procedure used in \citet{fabbri2021summevalreevaluatingsummarizationevaluation}. We observe strong alignment between annotators as indicated by the inter-annotator agreements presented in Table \ref{alphas}, measured via the Krippendorff’s alpha coefficient. Given the demanding nature of the annotation task and cost of human expert annotations, since the agreement is high additional annotations were deemed not necessary.

\begin{table}[t]
\centering \footnotesize\setlength{\tabcolsep}{5pt}
\begin{tabular}{ccc}\mediumhline
& Multi-XScience & QMDSCNN\\\hline
Krippendorff $\alpha$ & 0.91 & 0.94 \\ \mediumhline
\end{tabular} 
\caption{Inter-annotator agreements for human scores on summarization performance across six base models.}\label{alphas}
\end{table}

Based on computed correlations with scores from the annotators on 500 samples per dataset, we find GPT-3.5-Turbo and Gemini-1.5-Flash to be most human-aligned for evaluation of model performances on Multi-XScience and QMDSCNN, respectively, and therefore use these for all reported evaluations. The correlations are displayed in Table \ref{corrs}. These values are comparable with the human-LLM score correlations observed in leading LLM evaluation works such as G-Eval \citep{liu-etal-2023-g}, from which we adapted our LLM evaluation prompts.

\begin{table}[t]
\centering \footnotesize\setlength{\tabcolsep}{5pt}
\begin{tabular}{ccc}\mediumhline
& Multi-XScience & QMDSCNN\\\hline
Pearson's $r$ & 0.54	&0.58
 \\ \mediumhline
\end{tabular} 
\caption{Correlations between LLM and average human scores of summarization performance across six base models. GPT-3.5-Turbo is used for evaluation of Multi-XScience performance and Gemini-1.5-Flash is used for evaluation of QMDSCNN performance.}\label{corrs}
\end{table}

\section{Meta-Evaluation on \methodrm}  \label{metaevaldeets}

To validate the reliability of using \methodrm as an evaluator for MD instruction data, we conducted a meta-evaluation study to compare \methodrm with humans. 
Here we provide additional details regarding the prompt and setup used in the meta-evaluation study. We utilized two annotators (graduate students in NLP working directly with LLMs) to obtain fine-grained scores from 1-5 according to the six criteria defined in \method on 100 randomly sampled \method-generated candidate instructions. Each candidate was generated using a different prompt template over distinct news article clusters from NewSHead. Annotators were informed of the purpose, aims, and intended use of the study and annotations, and informed consent was collected prior to their performing the task. No compensation was provided given the small-scale nature of the task. The meta-evaluation task was formulated using the direction shown in Figure \ref{metaevalprompt}, and ratings were collected via a Google form. 

\begin{figure*}
\centering
 \begin{tcolorbox}[colframe=black, colback=gray!5, boxrule=0.5pt, arc=2mm, width=\textwidth, left=1mm, right=1mm, top=1mm, bottom=1mm, title=Meta-Evaluation Annotation Task Directions]\small
\textbf{Task Description}\\
In this task, you will evaluate the quality of instruction-answer pairs generated by a large language model (LLM). Each pair is based on a set of related news articles, with the instruction potentially being a question, command, or descriptive phrase.\\

Your goal is to assess how helpful each instruction-answer pair would be for instruction-tuning an LLM, specifically to improve its multi-document capabilities.\\

To this end, you will score each instruction-answer pair across six key criteria, which focus on both the pair's overall quality (criteria 1-3) and the pair’s effectiveness in handling multi-document content (criteria 4-6). While evaluating, prioritize the instruction over the answer, except where the criterion explicitly calls for focus on the answer (e.g., criterion \#2). But loosely keep in mind the quality of the answer, especially in cases where the answer’s accuracy or depth affects the overall value of the pair.\\

To correctly complete the task, please follow these steps:
\begin{itemize}
    \item Keep this document open on the side, such that this document and the Google Form for responses are both visible at once.
    \item Read the context documents, to be aware of the information contained in each article. Note that some samples have many documents or very long documents. It is not required to scrutinize the articles in meticulous detail; rather, the annotator should obtain an understanding of the content so as to make accurate judgments regarding the instruction-answer pair.
    \item Read the proposed instruction and answer.
    \item Rate the instruction-answer pair on a scale from 1 (worst) to 5 (best) according to the following criteria: relevance, coherence \& factuality, creativity, context integration, inter-document relationships, and complexity.
\end{itemize}

\textbf{Criteria Definitions}
\begin{itemize}
    \item \textbf{Relevance}: How well does the instruction align with the context provided by the documents? This involves assessing the pertinence of the instruction to the content and the degree to which it makes sense given the information in the documents.
    \item \textbf{Coherence \& Factuality}: Do the instruction and answer form a coherent, logical, and factually accurate pair? This involves checking if the answer directly addresses the instruction, if the level of detail in the answer is appropriate for the instruction, and if the reasoning behind the answer is sound and well-supported by the documents.
    \item \textbf{Creativity}: How creative or diverse is the instruction in terms of the type of question or task it involves? This involves assessing the question type (e.g., factual, inferential, analytical, summarization, etc.) and answer format (e.g., multiple-choice, open-ended, etc.).
    \item \textbf{Context Integration}: How well does the instruction leverage the context provided by multiple documents? This assesses whether the instruction involves synthesizing information from various documents to form a comprehensive response.
    \item \textbf{Inter-Document Relationships}: How well does the instruction encourage understanding relationships between different documents? This assesses whether the instruction requires comparing, contrasting, drawing connections, or identifying discrepancies between the documents.
    \item \textbf{Complexity}: Does the instruction appropriately and effectively challenge the answerer’s ability to understand and leverage information across multiple documents? This evaluates how well the instruction involves higher-order thinking skills like analysis, synthesis, and evaluation.
\end{itemize}
\end{tcolorbox}
\caption{G-Eval prompt for NLG evaluation of the fluency of a given summarization response.}
\label{metaevalprompt}
\end{figure*}

\begin{table}[h!]
\centering \scriptsize\setlength{\tabcolsep}{5pt}
\begin{tabular}{ccccc}\mediumhline
& Human & Human \& GPT & Human \& \methodrm \\\hline
Krippendorff $\alpha$ & 0.81 & 0.18  & 0.33 \\ \mediumhline
\end{tabular} 
\caption{IAA for human annotations, versus IAA and correlation for human and LLM annotations.}\label{metaevalresult}
\end{table}

The results of the meta-evaluation study are shown in Table \ref{metaevalresult}. We compute the inter-annotator agreement (IAA) among human annotators by calculating the Krippendorff’s alpha coefficient for each criterion and then taking the average over all 6 criteria. We also compute the IAA when GPT-3.5-Turbo and \methodrm are each separately considered as an additional annotator to demonstrate the value of \methodrm as a cost-effective, more human-aligned alternative to proprietary model scoring. Overall, we find that while GPT-3.5-Turbo and \methodrm have weak agreement with human annotators, \methodrm yields stronger alignment with humans versus GPT. This is consistent with demonstrated MD performance gains when using \methodrm-filtered instruction data as opposed to GPT-filtered data for IT. We note that since \methodrm is trained using data with target scores generated by GPT-4o, its alignment with human scoring could perhaps be improved by using human annotations of training data in future work.

\section{Full Experimental Results}  \label{fullresultsappendix}

We display full experimental results, previously abbreviated in Table \ref{maintable}, in Table \ref{apptable}.

\begin{table*}[h!]
\centering
\scriptsize\setlength{\tabcolsep}{1.35pt}
\begin{tabular}{@{}cc|l*{4}{c}|*{6}{c}|*{11}{c}|*{1}{c}@{}}
\toprule
&  & \multirow{2}{*}{\parbox[l]{1.2cm}{\centering Model / IT Data Setting}} & \multirow{2}{*}{HQA} & \multirow{2}{*}{WH} & \multirow{2}{*}{MXS} & \multirow{2}{*}{QC} & \multicolumn{6}{c}{SEAM} & \multicolumn{11}{c}{ZeroScrolls (ZS)} & \multirow{2}{*}{\centering Avg} \\
\cmidrule(lr){8-13} \cmidrule(l){14-24}
& & & & & & & ECB+ & MN & MSQ & OA&SC & Avg & GvRp & SSFD & QMsm & NrQA & Qspr & QLTY & SQLT & MuSQ & SpDg & BkSS & ZSS & \\
\midrule
\multirow{8}{*}{\rotatebox[origin=c]{90}{\parbox[c]{3.1cm}{\centering \textsc{FlanT5}}}} & \multirow{4}{*}{\rotatebox[origin=c]{90}{\parbox[c]{1.5cm}{\centering Base}}}  
& \cellcolor{BaselineColor} PRIMERA &\cellcolor{BaselineColor} 0.4 &\cellcolor{BaselineColor} 0.5 &\cellcolor{BaselineColor} 70.7	&\cellcolor{BaselineColor}	24.2 &\cellcolor{BaselineColor}7.1	&\cellcolor{BaselineColor}7.7	&\cellcolor{BaselineColor}0.5	&\cellcolor{BaselineColor}3.9&\cellcolor{BaselineColor}10.4	&\cellcolor{BaselineColor}5.9&\cellcolor{BaselineColor} 6.9	&\cellcolor{BaselineColor}	4.6	&\cellcolor{BaselineColor}	3.5	&\cellcolor{BaselineColor}	0.6	&\cellcolor{BaselineColor}	1.5	&\cellcolor{BaselineColor}	13.4	&\cellcolor{BaselineColor}	9.7	&\cellcolor{BaselineColor}	0.2	&\cellcolor{BaselineColor}	1.9	&\cellcolor{BaselineColor}	0.0	&\cellcolor{BaselineColor}	4.2	&\cellcolor{BaselineColor}8.8\\
& & \cellcolor{BaselineColor} \textsc{QAMDen} &\cellcolor{BaselineColor} 1.8 &\cellcolor{BaselineColor} 1.9 &\cellcolor{BaselineColor} 63.6	&\cellcolor{BaselineColor}	27.1 &\cellcolor{BaselineColor} 0.0	&\cellcolor{BaselineColor}0.4	&\cellcolor{BaselineColor}0.0	&\cellcolor{BaselineColor}1.0&\cellcolor{BaselineColor}3.1	&\cellcolor{BaselineColor}0.9 &\cellcolor{BaselineColor} 6.0	&\cellcolor{BaselineColor}	6.2	&\cellcolor{BaselineColor}	5.7	&\cellcolor{BaselineColor}	2.4	&\cellcolor{BaselineColor}	6.1	&\cellcolor{BaselineColor}	3.4	&\cellcolor{BaselineColor}	7.1	&\cellcolor{BaselineColor}	0.9	&\cellcolor{BaselineColor}	0.1	&\cellcolor{BaselineColor}	0.0	&\cellcolor{BaselineColor}	3.8	&\cellcolor{BaselineColor}7.2\\
& & None & 4.4 & 45.1 & 38.7	&	48.0 & 0.0	& 5.8	& 0.2	&2.6& 0.1	& 1.7& 5.2	&	5.4	&	9.9	&	16.7	&	14.2	&	48.2	&	4.8	&	26.9	&	0.0	&	0.0	&	13.1 & 14.5\\
& & Unfiltered & 31.7 & 47.5 & 89.8	&	52.2 & 0.0	&6.7&	0.1&2.7&	0.2&	1.9 &5.3	&	8.6	&	12.4	&	15.5	&	28.7	&	46.6	&	6.2	&	30.0	&	54.3	&	2.4	&	21.0 & 23.2\\
& & GPT-Filtered & 44.1 & 46.0 & 92.4	&	54.2 & 0.0&	6.0	&0.2&2.7&	0.0&	1.8& 6.1	&	9.2	&	13.4	&	16.4	&	25.8	&	49.6	&	6.8	&	27.8	&	54.4	&	3.6	&	21.3 &24.1\\
& & \cellcolor{OurColor} \methodrm &\cellcolor{OurColor} \textbf{47.3} &\cellcolor{OurColor} \textbf{48.3} &\cellcolor{OurColor} \textbf{93.8}	&\cellcolor{OurColor}	\textbf{57.3} &\cellcolor{OurColor} 0.0&\cellcolor{OurColor}	7.5	&\cellcolor{OurColor}0.3&\cellcolor{OurColor}2.7	&\cellcolor{OurColor}0.0	&\cellcolor{OurColor}\textbf{2.1} &\cellcolor{OurColor} 6.6	&\cellcolor{OurColor}	9.2	&\cellcolor{OurColor}	14.5	&\cellcolor{OurColor}	16.1	&\cellcolor{OurColor}	34.6	&\cellcolor{OurColor}	48.2	&\cellcolor{OurColor}	8.1	&\cellcolor{OurColor}	31.2	&\cellcolor{OurColor}	54.4	&\cellcolor{OurColor}	3.4	&\cellcolor{OurColor}	\textbf{22.6}	&\cellcolor{OurColor}\textbf{25.4}\\
\cmidrule[0.7pt]{2-25}
& \multirow{4}{*}{\rotatebox[origin=c]{90}{\parbox[c]{1cm}{\centering Large}}} 
& None & 24.4 & 54.6 & 70.9	&	61.8 & 1.1	&7.9	&0.1	&3.1&0.3	&2.5 &19.5	&13.5&	15.5&	9.4&	24.4	&31.0	&21.1	&11.8&	44.4&	41.1	&23.2&	24.0\\
& & Unfiltered & 46.9 & 62.9 & 91.1	&	64.7 & 0.7&	8.0	&0.0&3.8	&0.1	&2.5 & 6.6	&	10.4	&	12.1	&	19.7	&	37.6	&	61.8	&	9.3	&	35.1	&	48.7	&	0.1	&	24.2	&27.3\\
& & GPT-Filtered & 46.3 & 65.1 & 91.8	&	64.0 & 0.5&	8.3	&0.0&3.9&	0.1	&2.6 & 7.2	&	9.9	&	11.9	&	19.8	&	39.1	&	63.4	&	8.3	&	33.8	&	48.8	&	0.0	&	24.2	&27.5\\
& & \cellcolor{OurColor} \methodrm &\cellcolor{OurColor} \textbf{49.6} &\cellcolor{OurColor} \textbf{66.1} &\cellcolor{OurColor} \textbf{93.1}	&\cellcolor{OurColor}	\textbf{66.0} &\cellcolor{OurColor} 1.2&\cellcolor{OurColor}	9.1	&\cellcolor{OurColor}0.1	&\cellcolor{OurColor}4.2&\cellcolor{OurColor}0.0&\cellcolor{OurColor}	\textbf{2.9} &\cellcolor{OurColor} 7.5	&\cellcolor{OurColor}	10.1	&\cellcolor{OurColor}	12.8	&\cellcolor{OurColor}	20.0	&\cellcolor{OurColor}	43.2	&\cellcolor{OurColor}	64.0	&\cellcolor{OurColor}	10.7	&\cellcolor{OurColor}	35.4	&\cellcolor{OurColor}	48.6	&\cellcolor{OurColor}	0.1	&\cellcolor{OurColor}	\textbf{25.3}	&\cellcolor{OurColor}\textbf{28.5}\\
\cmidrule[0.7pt]{1-25}
\multirow{8}{*}{\rotatebox[origin=l]{90}{\parbox[c]{2.8cm}{\centering \textsc{Qwen2-Ins}}}} & \multirow{4}{*}{\rotatebox[origin=c]{90}{\parbox[c]{1cm}{\centering 1.5B}}} 
& None & 21.5 & 17.8 & 93.3	&	73.3 & 9.4	& 10.6& 	0.5	& 5.0&13.0	& 7.7 & 17.8	&	11.4	&	15.0	&	13.3	&	33.8	&	48.2	&	16.7	&	14.5	&	58.7	&	10.5	&	24.0	&25.5\\
& & Unfiltered & 32.9 & 30.5 & 94.2	&	79.3 &15.7	& 12.0& 	0.4	& 5.0&16.8	& 10.0 &14.1	&	9.8	&	15.3	&	15.7	&	40.3	&	48.0	&	15.7	&	16.1	&	53.8	&	10.1	&	23.9	&27.7\\
& & GPT-Filtered & 33.3 & 32.9 & 94.2	&	81.3 & 16.1	&12.0	&0.5	&5.1&15.9&	9.9 &17.9	&	11.3	&	14.5	&	13.8	&	35.9	&	49.2	&	16.5	&	16.0	&	56.9	&	10.1	&	24.2	&28.1\\
& & \cellcolor{OurColor} \methodrm &\cellcolor{OurColor} \textbf{37.7} &\cellcolor{OurColor} \textbf{34.8} &\cellcolor{OurColor} \textbf{94.4}	&\cellcolor{OurColor}	\textbf{82.9} &\cellcolor{OurColor} 16.4	&\cellcolor{OurColor} 12.0	&\cellcolor{OurColor} 0.6&\cellcolor{OurColor} 5.9	&\cellcolor{OurColor}18.1	&\cellcolor{OurColor} \textbf{10.6} &\cellcolor{OurColor}19.8	&\cellcolor{OurColor}	12.0	&\cellcolor{OurColor}	14.9	&\cellcolor{OurColor}	13.1	&\cellcolor{OurColor}	38.0	&\cellcolor{OurColor}	55.0	&\cellcolor{OurColor}	16.4	&\cellcolor{OurColor}	19.7	&\cellcolor{OurColor}	54.9	&\cellcolor{OurColor}	12.5	&\cellcolor{OurColor}	\textbf{25.6}	&\cellcolor{OurColor}\textbf{29.4}\\
\cmidrule[0.7pt]{2-25}
& \multirow{4}{*}{\rotatebox[origin=c]{90}{\parbox[c]{1.3cm}{\centering 7B}}} 
& \cellcolor{BaselineColor} LongAlign-7B &\cellcolor{BaselineColor} 10.4 &\cellcolor{BaselineColor} 14.3 &\cellcolor{BaselineColor} 92.2	&\cellcolor{BaselineColor}	83.3 &\cellcolor{BaselineColor} 11.5&\cellcolor{BaselineColor}	16.5	&\cellcolor{BaselineColor}0.0&\cellcolor{BaselineColor}4.1	&\cellcolor{BaselineColor}16.8	&\cellcolor{BaselineColor}9.8 &\cellcolor{BaselineColor}19.5	&\cellcolor{BaselineColor}	13.5	&\cellcolor{BaselineColor}	15.5	&\cellcolor{BaselineColor}	9.4	&\cellcolor{BaselineColor}	24.4	&\cellcolor{BaselineColor}	31.0	&\cellcolor{BaselineColor}	21.1	&\cellcolor{BaselineColor}	11.8	&\cellcolor{BaselineColor}	44.4	&\cellcolor{BaselineColor}	41.1	&\cellcolor{BaselineColor}	23.2	&\cellcolor{BaselineColor}25.3\\
& & None & 30.5 & 39.6 & 95.6	&	79.3 & 5.0	& 11.9	& 0.5	&6.4 &13.1	& 7.4 & 21.1	&	12.2	&	15.0	&	13.2	&	33.8	&	72.8	&	19.0	&	29.4	&	16.3	&	6.5	&	23.9	&27.4\\
& & Unfiltered & 40.5 & 43.3 & 94.7	&	84.2 & 8.0	&15.3	&0.5	&6.6&11.9&	8.5& 14.4	&	11.9	&	16.5	&	17.5	&	42.7	&	70.2	&	17.4	&	29.6	&	35.1	&	7.4	&	26.3	&29.9\\
& & GPT-Filtered & 42.0 & 44.0 & 94.7	&	85.3 & 8.7	&15.3	&0.9	&6.6&12.1	&8.7 & 11.5	&	12.0	&	16.8	&	18.1	&	43.1	&	67.4	&	15.9	&	28.4	&	54.1	&	19.6	&	28.7	&31.4\\
& & \cellcolor{OurColor} \methodrm &\cellcolor{OurColor} \textbf{44.7} &\cellcolor{OurColor} \textbf{46.0} &\cellcolor{OurColor} \textbf{95.1}	&\cellcolor{OurColor}	\textbf{87.3} &\cellcolor{OurColor} 13.8	&\cellcolor{OurColor}15.4	&\cellcolor{OurColor}0.6	&\cellcolor{OurColor}6.7&\cellcolor{OurColor}14.9&\cellcolor{OurColor}	\textbf{10.3} &\cellcolor{OurColor} 25.8	&\cellcolor{OurColor}	13.2	&\cellcolor{OurColor}	15.3	&\cellcolor{OurColor}	14.4	&\cellcolor{OurColor}	45.8	&\cellcolor{OurColor}	74.8	&\cellcolor{OurColor}	19.7	&\cellcolor{OurColor}	30.8	&\cellcolor{OurColor}	26.0	&\cellcolor{OurColor}	31.7	&\cellcolor{OurColor}	\textbf{29.8}	&\cellcolor{OurColor}\textbf{32.7}\\
\cmidrule[0.7pt]{1-25}
\multirow{6}{*}{\rotatebox[origin=l]{90}{\parbox[c]{3cm}{\centering \textsc{Llama3.1-Ins}}}} & \multirow{4}{*}{\rotatebox[origin=c]{90}{\parbox[c]{1.3cm}{\centering 8B}}} 
& \cellcolor{BaselineColor} ProLong-8B &\cellcolor{BaselineColor} 43.6 &\cellcolor{BaselineColor} 34.3 &\cellcolor{BaselineColor} 85.8	&\cellcolor{BaselineColor}	54.7 &\cellcolor{BaselineColor}9.5 &\cellcolor{BaselineColor}	17.4	 &\cellcolor{BaselineColor}0.4	 &\cellcolor{BaselineColor}10.7	 &\cellcolor{BaselineColor}18.0	 &\cellcolor{BaselineColor}11.2  &\cellcolor{BaselineColor}23.8	&\cellcolor{BaselineColor}	16	&\cellcolor{BaselineColor}	17.5	&\cellcolor{BaselineColor}	23.5	&\cellcolor{BaselineColor}	39.9	&\cellcolor{BaselineColor}	70.6	&\cellcolor{BaselineColor}	22.6	&\cellcolor{BaselineColor}	19.2	&\cellcolor{BaselineColor}	54.6	&\cellcolor{BaselineColor}	48.5	&\cellcolor{BaselineColor}	33.6	&\cellcolor{BaselineColor}	32.1\\
& & None & 35.5 & 27.1 & 95.1	&	65.3 & 10.5&	15.0	&0.6&7.6&	17.4	&10.2 &23.7	&	5.8	&	5.2	&	10.5	&	4.6	&	75.6	&	13.3	&	0.7	&	47.2	&	0.3	&	18.7	& 24.3\\
& & Unfiltered & 37.6 & 34.4 & 84.7	&	90.4& 15.3&	16.3	&0.5&7.8&	18.9&	11.8 & 20.2	&	14.1	&	17.4	&	21.9	&	52.3	&	50.0	&	19.4	&	25.3	&	32.6	&	32.1	&	28.6	&31.1\\
& & GPT-Filtered & 38.0 & 42.3 & 95.3	&	87.8& 8.7	&16.0	&0.6	&7.8&18.3	&10.3& 20.3	&	14.2	&	17.5	&	22.2	&	53.6	&	59.8	&	19.5	&	26.2	&	36.1	&	26.4	&	29.6	&32.1\\
& & \cellcolor{OurColor} \methodrm &\cellcolor{OurColor} \textbf{44.7} &\cellcolor{OurColor} \textbf{43.7} &\cellcolor{OurColor} \textbf{95.3}	&\cellcolor{OurColor}	\textbf{93.8} &\cellcolor{OurColor} 16.3	&\cellcolor{OurColor}16.4&\cellcolor{OurColor}	0.6	&\cellcolor{OurColor}7.9&\cellcolor{OurColor}18.5	&\cellcolor{OurColor}\textbf{11.9}&\cellcolor{OurColor} 19.9	&\cellcolor{OurColor}	14.5	&\cellcolor{OurColor}	17.6	&\cellcolor{OurColor}	22.6	&\cellcolor{OurColor}	52.2	&\cellcolor{OurColor}	68.2	&\cellcolor{OurColor}	19.2	&\cellcolor{OurColor}	27.5	&\cellcolor{OurColor}	34.3	&\cellcolor{OurColor}	33.4	&\cellcolor{OurColor}	\textbf{30.9}	&\cellcolor{OurColor}\textbf{34.0}\\
\cmidrule[0.7pt]{2-25}
& \multirow{2}{*}{\rotatebox[origin=c]{90}{\parbox[c]{1.3cm}{\centering 70B}}} 
& \cellcolor{BaselineColor} Jamba 1.5 Mini &\cellcolor{BaselineColor} 47.5 &\cellcolor{BaselineColor} 41.8 &\cellcolor{BaselineColor}94.2	&\cellcolor{BaselineColor}	87.1 &\cellcolor{BaselineColor} 20.1	&\cellcolor{BaselineColor}14.2	&\cellcolor{BaselineColor}0.0	&\cellcolor{BaselineColor}5.3&\cellcolor{BaselineColor}20.4	&\cellcolor{BaselineColor}12.0 &\cellcolor{BaselineColor} 21.1	&\cellcolor{BaselineColor}	15.1	&\cellcolor{BaselineColor}	17.3	&\cellcolor{BaselineColor}	22.2	&\cellcolor{BaselineColor}	47.3	&\cellcolor{BaselineColor}	84.4	&\cellcolor{BaselineColor}	20	&\cellcolor{BaselineColor}	28.1	&\cellcolor{BaselineColor}	59.1	&\cellcolor{BaselineColor}	26	&\cellcolor{BaselineColor}	34.1	&\cellcolor{BaselineColor}	35.3\\
 & & None & 53.9 & 38.1 & 95.1	&	88.2 & 25.1	& 21.9& 	0.6	&6.1& 11.5	& 13.0 & 21.9	&	14.9	&	18.1	&	25.5	&	45.5	&	66.7	&	23.4	&	42.5	&	61.4	&	43.8	&	36.4	&37.1\\
& & Unfiltered & 55.9 & 40.6 & 88.7	&	70.4 &3.6	&21.9&	0.6	&6.3	&11.2&	8.7 & 20.5	&	14.8	&	17	&	32.4	&	49.7	&	77.4	&	18	&	33.1	&	62.8	&	23.4	&	34.9	&	34.1\\
& & GPT-Filtered & 57.4 & 41.2 & 88.2	&	74.9 & 4.9	&22.0	&0.7&	6.4	&10.7&	8.9& 21.5	&	15.3	&	18.3	&	30.1	&	50	&	83.8	&	19.3	&	32.7	&	65.2	&	38.7	&	37.5	&	35.9\\
& & \cellcolor{OurColor} \methodrm &\cellcolor{OurColor} \textbf{58.4 }&\cellcolor{OurColor} \textbf{45.5} &\cellcolor{OurColor} \textbf{95.1}	&\cellcolor{OurColor}	\textbf{88.7} &\cellcolor{OurColor} 25.2	&\cellcolor{OurColor} 22.1&\cellcolor{OurColor} 	0.7	&\cellcolor{OurColor}6.7&\cellcolor{OurColor} 12.0	&\cellcolor{OurColor}\textbf{ 13.3}&\cellcolor{OurColor} 21.5	&\cellcolor{OurColor}	15.4	&\cellcolor{OurColor}	18.3	&\cellcolor{OurColor}	29.8	&\cellcolor{OurColor}	50.7	&\cellcolor{OurColor}	83.4	&\cellcolor{OurColor}	19.4	&\cellcolor{OurColor}	33.6	&\cellcolor{OurColor}	66.0	&\cellcolor{OurColor}	38.7	&\cellcolor{OurColor}	\textbf{37.7}	&\cellcolor{OurColor}\textbf{38.5}\\
\cmidrule[0.7pt]{1-25}
& & GPT-4o & 57.5 & 50.0 & \textbf{100.0}	&	\textbf{94.4} & 8.9	 &18.6 &	0.3 &14.2&	28.9	 &14.1 & 25.7	&	14.8	&	15.9	&	35.3	&	58.8	&	88.6	&	20.7	&	49.2	&	83.5	&	5.6	&	\textbf{39.8}	&\textbf{40.6}\\
& & Gemini 1.5 Pro & \textbf{66.6} & \textbf{55.6} & 93.8	&	79.8 & 21.4&	17.8	&0.6	&15.1&30.1	&\textbf{17.0}& 22.1	&14.3&	14.3	&53.9&	36.1&	85.7	&18.1&	36.1	&39.4&	0.0	&32.0  &{36.9}\\
\bottomrule
\end{tabular}
\caption{Full evaluation of MDCure'd models versus baselines on 6 benchmarks in the zero-shot prompting setting. The rightmost ``Avg'' column reports the average of individual dataset scores. Dataset abbreviations are described in App. \ref{abbreviations}. Rows specified by ``\methodrm'' refer to our full \method pipeline applied to the corresponding base model and size. Size-comparable baselines are highlighted in \hlcolorone{blue} and results of our final method is highlighted in \hlcolortwo{yellow}. Bold numbers show best performance in each group.}\label{apptable}
\vspace{-2.5mm}
\end{table*}
\end{document}